\documentclass[10pt,twocolumn,letterpaper]{article}

\usepackage[pagenumbers]{cvpr} %

\usepackage{microtype}

\renewcommand{\paragraph}[1]{\vspace{.5em}\noindent\textbf{#1.}}

\usepackage{soul}
\setuldepth{foobar}

\usepackage{colortbl}
\usepackage{siunitx}
\usepackage{comment}

\usepackage[most]{tcolorbox}
\usepackage{xurl}
\usepackage{xparse}

\usepackage{booktabs}
\usepackage{arydshln} %
\usepackage{wrapfig}
\usepackage{refcount}
\usepackage[dvipsnames]{xcolor}
\newcommand{\dpos}[1]{\textcolor{ForestGreen!70!black}{#1}}
\newcommand{\dneg}[1]{\textcolor{BrickRed!70!black}{#1}}

\NewDocumentCommand{\projectsite}{o}{%
  \IfNoValueTF{#1}%
    {\href{https://flat-pack-bench.github.io}{\nolinkurl{flat-pack-bench.github.io}}}%
    {\href{https://flat-pack-bench.github.io/#1}{\nolinkurl{flat-pack-bench.github.io/#1}}}%
}

\usepackage{supertabular}

\definecolor{rankA}{HTML}{1B5E20} %
\definecolor{rankB}{HTML}{43A047} %
\definecolor{rankC}{HTML}{A5D6A7} %
\definecolor{rankD}{HTML}{FFFFFF} %
\definecolor{sectionblue}{HTML}{E3F2FD} %
\newcommand{\sectiondivider}[1]{%
  \noalign{\vskip-1pt}\specialrule{1.5pt}{0pt}{0pt}\noalign{\vskip-1pt}
  \rowcolor{sectionblue}\multicolumn{7}{l}{\textit{#1}}\\
  \hdashline
}

\newcommand{\sectiondividertwo}[1]{%
  \noalign{\vskip-1pt}\specialrule{1.5pt}{0pt}{0pt}\noalign{\vskip-1pt}
  \rowcolor{sectionblue}\multicolumn{11}{l}{\textit{#1}}\\
  \hdashline
}

\newcommand{\rankcell}[1]{%
  \begingroup
  \def\r{#1}%
  \ifnum\r<2 \cellcolor{rankA}{\color{white}\bfseries #1}%
  \else\ifnum\r<3 \cellcolor{rankB}{\color{white} #1}%
  \else\ifnum\r<4 \cellcolor{rankC}{#1}%
  \else \cellcolor{rankD}{#1}%
  \fi\fi\fi
  \endgroup
}

\definecolor{darkgray}{gray}{0.75} %
\newcommand{\best}[1]{\cellcolor{darkgray}{\bfseries #1}}
\definecolor{lightgray}{gray}{0.25} %
\newcommand{\bestopen}[1]{{\underline{#1}}}
\newcommand{\highlight}[2]{%
  \raisebox{0pt}[0pt][0pt]{\colorbox{#1}{#2}}%
}

\definecolor{boxborder}{HTML}{004444} %
\definecolor{boxfill}{RGB}{240,240,240}
\newtcolorbox{emphquote}{
  colback=boxfill,
  colframe=teal!60!black,   %
  boxrule=0.5pt,            %
  arc=4pt,                  %
  left=8pt, right=8pt, top=6pt, bottom=6pt, %
  width=\linewidth,
}

\newcommand{\flatpack}{\textsc{Flat-Pack Bench}}
\definecolor{cvprblue}{rgb}{0.21,0.49,0.74}
\usepackage[pagebackref,breaklinks,colorlinks,allcolors=cvprblue]{hyperref}

\definecolor{lightgray}{gray}{0.9}
\definecolor{lightgreen}{RGB}{210, 255, 210}
\definecolor{midgreen}{RGB}{150, 235, 150}
\definecolor{darkgreen}{RGB}{100, 200, 100}

\def\paperID{}
\def\confName{CVPR}
\def\confYear{2026}

\title{\flatpack: Evaluating Spatio-Temporal Understanding in Large Vision-Language Models through Furniture Assembly}

\author{
\textbf{Aditya Chetan$^{1}$\thanks{Correspondence: {\tt achetan@cs.cornell.edu}}} \hspace{12pt}
\textbf{Eric Cai$^{1}$} \hspace{12pt}
\textbf{Peeyush Kushwaha}\thanks{Independent researcher} \hspace{12pt}
\textbf{Bharath Raj Nagoor Kani$^{1}$} \hspace{12pt}
\\
\textbf{Utkarsh Mall$^{3}$} \hspace{12pt}
\textbf{Qianqian Wang$^{4}$} \hspace{12pt}
\textbf{Noah Snavely$^{1,2}$} \hspace{12pt}
\textbf{Bharath Hariharan$^{1}$} \hspace{12pt}
\\
$^{1}$Cornell University \
${^2}$Cornell Tech \
${^3}$MBZUAI \
${^4}$UC Berkeley \\
\projectsite}

\begin{document}
\newcommand{\bolt}{\textsc{BOLT}}

\newcommand{\fpb}{\textsc{FPB}}
\newcommand{\vsibench}{\textsc{VSI-Bench}}
\newcommand{\imaw}{IMaW}
\newcommand{\mating}{\textsc{Mate}}
\newcommand{\ordering}{\textsc{TOrd}}
\newcommand{\localization}{\textsc{TLoc}}
\newcommand{\track}{\textsc{Track}}

\newcommand{\aditya}[1]{{\textcolor{red}{AC\@: {#1}}}}
\newcommand{\peeyush}[1]{\comment{\textcolor{cyan}{PK\@: {#1}}}}
\newcommand{\bharath}[1]{\textcolor{magenta}{BH\@: {#1}}}
\newcommand{\noah}[1]{\comment{\textcolor{orange}{NS\@: {#1}}}}
\newcommand{\utkarsh}[1]{\textcolor{red}{UM\@: {#1}}}

\newcommand{\compactpara}[1]
{\noindent \textbf{#1}}

\newcommand{\tva}{\textsc{TVA}}
\newcommand{\dashfill}{\leavevmode\leaders\hbox{-}\hfill\kern0pt}
\definecolor{myorange}{RGB}{200, 90, 0}
\definecolor{mydarkblue}{RGB}{0, 60, 125}     %
\definecolor{mymagenta}{RGB}{200, 0, 150}     %

\newcommand{\orange}[1]{\textcolor{myorange}{#1}}
\newcommand{\darkblue}[1]{\textcolor{mydarkblue}{#1}}
\newcommand{\magenta}[1]{\textcolor{mymagenta}{#1}}

\maketitle
\begin{abstract}
The emergence of Large Vision-Language Models (LVLMs) has significantly advanced video understanding capabilities. 
However, existing benchmarks focus predominantly on coarse-grained tasks such as action segmentation, classification, captioning, and retrieval. 
Furthermore, these benchmarks often rely on entities that can be easily identified verbally, like household objects, animals, human subjects, etc., limiting their applicability to complex, in-the-wild video scenarios. 
But, many applications such as furniture assembly, cooking, etc., require step-by-step fine-grained spatio-temporal understanding of the video, which is not sufficiently evaluated in current benchmarks.
To address this gap, we introduce \textnormal{\flatpack}, a novel benchmark centered on furniture assembly tasks. 
Our benchmark evaluates LVLMs on nuanced tasks, including temporal ordering of assembly actions, temporal localization of assembly state, understanding part mating, and tracking, using multiple-choice questions paired with visual prompts highlighting relevant parts as references for fine-grained questions. 
Our experiments reveal that state-of-the-art LVLMs struggle significantly with fine-grained spatio-temporal reasoning, highlighting their limitations in effectively leveraging temporal information from videos, limited tracking ability, and understanding of spatial interactions like physical contact.
\end{abstract}
\vspace{-1em}
    
\section{Introduction}
\label{sec:intro}

\begin{figure*}
    \centering
\includegraphics[width=0.95\linewidth]{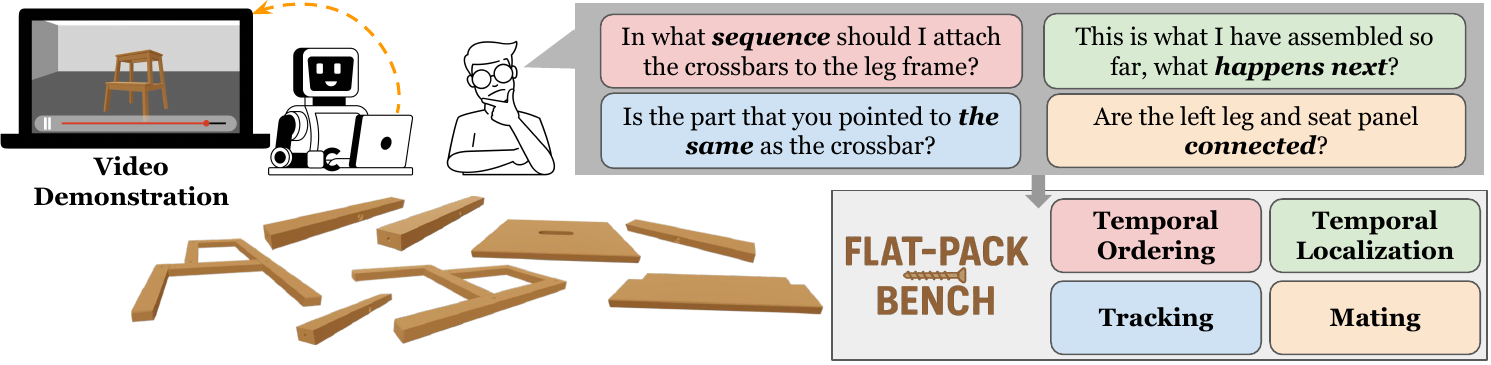}
\captionof{figure}{\textbf{Motivation for \flatpack}. For AI assistants to understand an assembly process through observation, they need to be adept at fine-grained spatio-temporal reasoning about the video. We propose \flatpack{} to evaluate Large Vision-Language Models on four such fine-grained video understanding tasks, namely -- Temporal Ordering, Temporal Localization, Tracking, and Mating.}
\label{fig:teaser}
\vspace{-1.6em}
\end{figure*}

Imagine an AI assistant that helps us out with complex but practical everyday tasks, like cooking a complicated recipe, repairing equipment, or assembling furniture.
We may want this AI assistant to watch an instructional video or demonstration of the task, then answer questions to help us out: what ingredients should I add next? Which pipes should connect? Which of these pieces should be screwed in first?
A natural way to build such an AI assistant is to use Large Vision-Language Models (LVLMs)~\cite{openai2025gpt5,geminiteam2025geminifamilyhighlycapable,zhu2025internvl3exploringadvancedtraining,bai2025qwen25vltechnicalreport,zhang2024llavavideo,li2024llavaonevision,zhang2024llavanextvideo} -- after all, these models exhibit broad understanding of images and videos and can interact in natural language. 
But are current LVLMs up to the task?

To answer this question, it is useful to walk through the 
required skills.
The first step of course is understanding the demonstration video. 
This itself presents significant challenges.
Simply understanding the overall objective (``What is being done in this video?") is not enough: the LVLM must understand what each individual step is, how to accomplish that step, and when to execute it, and then communicate this understanding with a human user.
Understanding how to execute individual steps requires the model to detect and localize 
particular objects (e.g., recipe ingredients or furniture parts) and recognize how they interact (e.g., detect when two parts are screwed together).
Understanding when to execute what step requires tracking the ingredients or assembly components over long demonstrations with multiple steps.
Finally, communicating this understanding in cluttered environments requires the model to understand not just text but also spatial references~\cite{yuan_videorefer_2025,zhou_strefer_2025}.

Existing benchmarks do not stress-test LVLMs on such challenges.
Many video QA benchmarks focus on short videos~\cite{li2024mvbenchcomprehensivemultimodalvideo,xiao_next-qa_2021}, or on coarse-grained questions about video themes that do not require temporal understanding or tracking~\cite{mangalam2023egoschema}.
Often the scenes are uncluttered~\cite{caba2015activitynet,zheng2024contphycontinuumphysicalconcept,wang2025compositional4ddynamicscenes} making object references unambiguous~\cite{cho2025PerceptionLM,yuan_videorefer_2025, zhou_strefer_2025}.
There is also limited focus on modeling object interactions that are ubiquitous in 
complex, everyday activities.

In this paper, we address this gap with a new video QA benchmark 
using the task of furniture assembly as a sandbox  (Figure~\ref{fig:teaser}).
Furniture assembly is a simplified microcosm of precisely the challenges we identify above: recognizing and tracking objects over multiple steps and detecting their interactions in a cluttered visual scene.
The simplicity of the domain (rigid parts that retain shape and identity throughout the video) allows us to precisely articulate the skills that current models lack.
Success in this domain is a 
prerequisite for more complex domains where object state might change (e.g., tomatoes getting cut or squished) and complex interactions between ingredients and parts might occur.

To build this benchmark, we 
augment prior furniture assembly datasets~\cite{liu2024ikea} with significant new annotations: 
segmentations of each part for spatial references, 
connections between individual parts, 
and a manually curated set of natural language multiple-choice questions.
The resulting benchmark, \flatpack\footnote{\textit{Flat-pack} furniture is a term used for ready-to-assemble furniture. Benches are commonly sold in flat-pack form, thus: \flatpack.}, tests multiple axes of temporal understanding, including whether models can track parts across frames, understand which parts connect to which, and determine the order in which different connections happen.

We test multiple proprietary and open 
LVLMs on our benchmark and find that even the best models struggle: OpenAI's latest GPT-5 model~\cite{openai2025gpt5} achieves an accuracy of \textbf{$\sim$38\%}, trailing far behind human performance of 94.18\%.
We look deeper into model performance, and show how models struggle with spatio-temporal reasoning tasks like tracking and contact detection required 
for the tasks in our benchmark.
We further explore an agentic approach~\cite{surismenon2023vipergpt} that uses standard, state-of-the-art vision models like SAM2~\cite{ravi2025sam} as tools, but find that the tools themselves struggle on these challenging videos.
Our benchmark suggests that, despite rapid progress, current LVLMs (and computer vision systems in general)
have limited capability to understand the temporal evolution of complex scenes.

\section{Related Work}

\paragraph{Video Understanding Benchmarks} Prior works on  Video Question Answering (VidQA) tend to focus on coarse-grained questions about high-level scene semantics~\cite{caba2015activitynet,goyal2017something,li2024mvbenchcomprehensivemultimodalvideo}. 
Recently there has been a surge of interest in benchmarks on physical scene understanding~\cite{zheng2024contphycontinuumphysicalconcept,wang2025compositional4ddynamicscenes} but these often focus on synthetic videos without clutter or occlusion.
On more real-world videos, some recent benchmarks seek to evaluate spatial intelligence~\cite{yang2024think}, but much of this work focuses on static scenes with camera motion (typically ignoring any dynamic objects)~\cite{li2025stibenchmllmsreadyprecise, yang2024think, zhou2025urbenchcomprehensivebenchmarkevaluating, zhang2025pointvisiontextdoes}.
VLM4D~\cite{zhou2025vlm4d} evaluates the relative motion understanding of LVLMs for dynamic scenes, but unlike our proposed benchmark, does not explore interactions between objects.
Temporal understanding abilities of LVLMs has also been studied with past works evaluating temporal sensitivity~\cite{xue2025seeing} and eliminating single-frame bias~\cite{zohar_apollo_2024} in existing benchmarks.
Our benchmark is also related to these works, but we focus on the fine-grained temporal evolution of the scene.
Interactive video-based guidance benchmarks~\cite{whattosay2024neurips,bao-etal-2023-foundation,wang2023holoassist,livecook} are also related, but they tend to focus on short-range temporal context: what has immediately transpired and what should happen next.

Similar to our approach, LEGO-Puzzles evaluates LVLMs on multi-step assembly-based reasoning~\cite{tang2025lego}. 
However, they focus on a \textit{multi-image} setting, providing 2-3 images of relevant assembly steps as inputs to the model.
This simplifies the problem and is not representative of real-world demonstrations which may have long demonstrations with multiple frames, as represented by 
 our benchmark: the model must decide what frames it should focus on.
Finally, recent benchmarks have also explored long videos and fine-grained questions~\cite{longvideobench, mangalam2023egoschema}, but often 
they focus on simple uncluttered scenes.
Cluttered scenes, like those in our benchmark, introduce additional challenges where the model has to disambiguate similar-looking parts, and understand not just textual but spatial references.

\paragraph{Regional Understanding in LVLMs}
Recent work has explored tasks where the LVLM must track a segmented object through a video and reason about the track~\cite{yuan_videorefer_2025,zhou_strefer_2025}.
However, the objects in these datasets are small in number (typically one or two per video) and easily tracked.
In contrast, our benchmark features manually curated and verified segmentations and questions involving complex interactions among multiple, visually similar parts—making it substantially more challenging, as it requires more difficult tracking.
Another related model–benchmark pair, PerceptionLM and PLM-VideoBench~\cite{cho2025PerceptionLM}, assume tracking to be solved, with the model provided full-video tracks, whereas our benchmark requires the model to track the highlighted regions specified in specific frames.
A related task is Spatio-Temporal Visual Grounding (STVG)~\cite{zhang2020does,tang2021human, munasinghe_videoglamm_2025, sun_sama_2025}, where models are given a video and a textual description of an object, and the model must segment the referred object.
In contrast to this problem, we do not expect segmentation masks as an output, though we do demonstrate the performance of an agent that uses segmentation/tracking as a tool.

\section{\flatpack}
\label{sec:dataset}
\begin{figure*}
    \centering
    \includegraphics[width=0.95\linewidth]{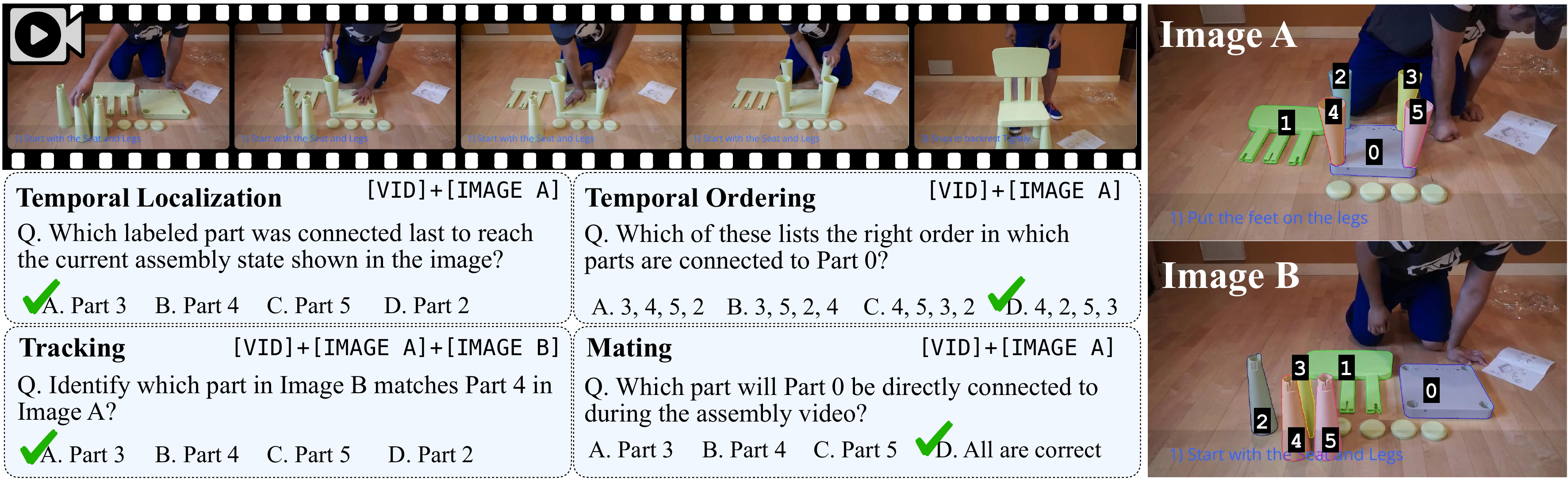}
    \caption{\textbf{Snapshot of \flatpack.} Each question consists of an assembly video (top row), one or two visual prompts (Images A, B), and a multiple-choice question. The corresponding visual inputs are shown within each question box. Videos are sourced from the internet and may include artifacts like overlaid text. For clarity, part labels are enlarged, as the visual prompts are shown at reduced scale.
    }
    \vspace{-1.5em}
    \label{fig:snapshot}
\end{figure*}

\begin{table}[t]
\centering
\caption{\textbf{Dataset composition}: Shows the number of videos (\textbf{\#V}), questions (\textbf{\#Q}), and templates per category (\textbf{\#T}), along with average questions per video (\textbf{Q/V}), per template (\textbf{Q/T}), and unique templates per video (\textbf{uT/V}).}
\renewcommand{\arraystretch}{1.2}
\renewcommand{\tabcolsep}{1.2mm}
\resizebox{\linewidth}{!}{
\begin{tabular}{
    l
    S[table-format=3.0]
    S[table-format=3.0]
    S[table-format=3.2]
    S[table-format=2.2]
    S[table-format=2.0]
    S[table-format=3.2]
    S[table-format=1.2]
}
\toprule
\textbf{Category} & {\textbf{\#V}} & {\textbf{\#Q}} & {\textbf{Q(\%)}} & {\textbf{Q/V}} & {\textbf{\#T}} & {\textbf{Q/T}} & {\textbf{uT/V}} \\
\midrule
\textsc{TLoc} & 35 & 103 & 17.11 & 2.94 & 6 & 17.17 & 1.71 \\
\textsc{TOrd}     & 39 & 155 & 25.75 & 3.97 & 2 & 77.50 & 1.51 \\
\textsc{Mate}                & 21 &  87 & 14.45 & 4.14 & 3 & 29.00 & 1.48 \\
\textsc{Track}              & 43 & 257 & 42.69 & 5.98 & 2 & 128.50 & 1.37 \\
\midrule
\textbf{Total}        & 50 & 602 & 100.00 & 12.04 & 13 & 46.31 & 4.18 \\
\bottomrule
\end{tabular}}
\label{tab:databreakdown}
\vspace{-1.6em}
\end{table}

We now describe the process of creating the benchmark.

\paragraph{Data} We base our benchmark on the IKEA-Manuals-at-Work (\imaw{}) dataset~\cite{liu2024ikea}.
\imaw{} contains in-the-wild videos of people assembling IKEA furniture pieces.
The dataset also comes with a limited set of annotations: 3D models for the furniture and the furniture parts, and key frames that are annotated with segmentation masks, the 6DoF pose for each part, and the \emph{sub-assemblies} (a set of parts) that are being connected. 
The original assembly demonstration videos contain unnecessary segments consisting of text-only instruction cards, typically when the timestamps of consecutive key frames have a gap greater than one second.
We use this heuristic to trim such sections from the raw videos, resulting in the \textit{trimmed videos}.
Together with videos composed solely of key frames (\textit{key-frame videos}), these form the two types of assembly videos used in our evaluations.
Key-frame videos contain only the most salient frames, offering a concise view of the assembly process, but require manual curation and are therefore unrealistic.
Trimmed videos are more realistic, but often longer, stressing current model capabilities.
We evaluate both settings in our experiments.

\paragraph{Visual Prompts}
One challenge with constructing questions on furniture-assembly videos is unambiguously referring to furniture parts.
Text descriptions (like ``top railing") may be ambiguous in symmetric structures, and may prompt the model to rely on its common-sense knowledge of typical furniture rather than the provided visual input.
Instead, we propose to use \emph{visual prompts}~\cite{yang2023set}, and mark out each part on an image prompt using a label and a segmentation.
Thus, each question in our benchmark consists of an assembly video, 1-2 visual prompts (constructed from frames in the same video) and a multiple-choice question.
We use the part's label in the visual prompt to refer it in the question.

Visual prompts require segmentations.
However, we find the segmentation annotations in \imaw{} to be incomplete
(See App.~\ref{app:vp} for details).
We therefore augment \imaw{} with manually annotated part segmentations on 
343 frames from a subset of 50 videos.
These frames serve as our visual prompts.
Next, we construct the questions in the dataset.

\paragraph{Question Types}
To fully understand a furniture assembly, the model must identify which parts connect (\textit{mating}) and when they connect. To be able to infer which connections happen when, it should \textit{track} the individual parts through the video.
Thus, we propose the following question types:

\begin{itemize}
\item \mating{}  asks 
the model to determine 
whether two parts are connected in the final assembly.
\item \track{} provides two segmented frames with shuffled part IDs as visual prompts and requires the model to recover the correct correspondences using the video.
\item \ordering{} evaluates whether the model can infer the correct order of connection events.
\item \localization{} tests its ability to identify events immediately before or after the state shown in the visual prompt, assessing temporal localization and reasoning about nearby events.
\end{itemize}
For each of these question types, we create templates; some examples are shown in \cref{fig:snapshot}. 
Formulating these questions requires annotations of part–part connections in the video.
However, as noted earlier, \imaw{} provides only sub-assembly–level annotations.
As such, we augment each video with fine-grained part assembly annotations that specify which parts connect, to which other parts, and when.

\paragraph{Manual Question Curation}
With these annotations and question templates, we can auto-generate questions.
However, we found that auto-generation frequently produced questions that could be answered by ignoring the video and exploiting shortcuts.
For example, auto-generated \textit{mating} questions about parts already positioned for connection, or included distractor options with clearly distinct shapes or colors, enabling easy elimination (See App.~\ref{app:questioncuration} for examples).
To address this, we curated all questions manually using fixed templates.
Annotators were provided the full assembly video, segmentation-labeled frames for visual prompts, the question templates, and detailed guidelines for avoiding shortcuts based on static cues from the visual prompt.
The full list of templates is provided in App.~\ref{app:questioncuration}.

\paragraph{Final Benchmark} The final benchmark consists of 602 multiple-choice questions covering 50 different furniture assembly videos sourced from the \imaw{}~\cite{liu2024ikea} dataset.
\Cref{fig:snapshot} shows a snapshot
and \cref{tab:databreakdown} shows the dataset composition. For additional details about the benchmark, see App.~\ref{app:qualexamples}.
More examples from the benchmark can be viewed at: \projectsite[viewer/]

\section{Evaluation on \flatpack}
\label{sec:eval}
\subsection{Setup}
\vspace{-0.5em}
\paragraph{Benchmark Models}
We evaluated the following models:
\begin{itemize}
    \item \textbf{Proprietary}: Gemini 2.5/3.1~\cite{geminiteam2025geminifamilyhighlycapable} \& GPT-5~\cite{openai2025gpt5}. 
    \item \textbf{Open}: Video-LLaVA~\cite{lin2023video}, LLaVA-NeXT-Vid~\cite{zhang2024llavanextvideo}, LLaVA-OneVision~\cite{li2024llavaonevision}, LLaVA-Video~\cite{zhang2024llavavideo}, Qwen 2.5/Qwen 3-VL~\cite{bai2025qwen25vltechnicalreport,qwen3vl} and InternVL3~\cite{zhu2025internvl3exploringadvancedtraining}. We cover models with a range of design choices, from fine-tuning image-text models with video data~\cite{zhang2024llavanextvideo}, to special architectural modifications to handle videos~\cite{zhu2025internvl3exploringadvancedtraining,bai2025qwen25vltechnicalreport,li2024llavaonevision}. 
    \item \textbf{Specialized}: We also evaluate models tailored for specific capabilities relevant to our task: 
    ArrowRL~\cite{xue2025seeing} improves temporal sensitivity; 
    PerceptionLM~\cite{cho2025PerceptionLM} and VideoRefer~\cite{yuan_videorefer_2025} target fine-grained regional understanding; 
    and GenS~\cite{yao-etal-2025-generative} selects question-relevant frames in long videos for a base LVLM (Gemini 2.5 Pro in our case). 
\end{itemize}
We evaluate all the models under a zero-shot setting.
Following \vsibench~\cite{yang2024think}, we use greedy decoding for all models (except for GPT-5, which does not support it) to ensure reproducibility.

\medskip \noindent \textbf{Prompt Construction.} The input prompt consists of a video, 1-2 visual prompts, the question, and a fixed task instruction that describes the multi-modal input and output format to the model. We employ three different settings to supply the visual prompt to the model:
\begin{itemize}
    \item \textbf{\textit{Mixed-media Prompt}}: The visual prompt is supplied as an image separate from the assembly video.
    \item \textbf{\textit{Collage Prompt}}: Each frame of the video is a grid of images containing the visual prompt(s) (fixed in each frame) and the original video frame.
    \item \textbf{\textit{Concat Prompt}}: The visual prompts are concatenated to the video as the first 1 (or 2) frame(s).
\end{itemize}
The task instructions describe the prompt format to the model. Previous works~\cite{yao-etal-2025-generative} have also shown that for long videos, finding salient key-frames for answering the questions can be challenging. 
To evaluate the effect of key-frames, we also try both trimmed 
and key-frame videos 
for all prompt formats.
We evaluate all configurations across all models, except when restricted by architectural or cost constraints.
See App.~\ref{app:evaldetails} for more experimental details.

\medskip \noindent \textbf{Chance Baselines.}
We also report the accuracy of choosing an answer uniformly at random (\textit{Random Chance}) or picking the most common option per task (\textit{Frequency Chance}).

\medskip \noindent \textbf{Human Performance.} We also evaluated how humans perform on the benchmark. 
We recruited a group of participants consisting of Computer Science students ranging from undergraduate to doctoral levels. 
Each participant was shown the assembly video, the visual prompt, the multiple-choice question, and the task instruction and asked to select an answer. 
We collected 3 responses for each question and selected the final response using majority voting.
We also conducted a broader crowd-sourced study on a randomly-sampled subset of questions (see App.~\ref{app:results} for the details and results).

\medskip \noindent \textbf{Metric.} We use accuracy as our metric with a regex-based exact match to extract the answer from model responses.

\subsection{Results}

\begin{table}[t]
\centering
\caption{\textbf{Results on \flatpack.} \highlight{darkgray}{\textbf{Best model}} and \underline{best open model} are highlighted in each column.}
\renewcommand{\arraystretch}{1.2}
\renewcommand{\tabcolsep}{1.2mm}
\resizebox{\linewidth}{!}{
\begin{tabular}{lcccccc}
\toprule
\textbf{Model} & \textbf{Rank} & \textbf{Micro Avg.} & \textbf{\textsc{TOrd}} & \textbf{\textsc{TLoc}} & \textbf{\textsc{Track}} & \textbf{\textsc{Mate}} \\
\midrule
\textbf{\textit{Human Performance}} & -- & 94.18 & 93.54 & 93.20 & 93.77 & 97.70 \\
\sectiondivider{Chance Baselines}
Random Chance & -- & 26.41 & 25.00 & 25.00 & 25.49 & 33.33 \\
Frequency Chance & -- & 26.74 & 27.74 & 30.10 & 26.46 & 36.78 \\
\sectiondivider{Proprietary Models}
GPT-5 & \rankcell{1} & 37.71 & 40.65 & \best{53.40} & 25.68 & 49.43 \\
Gemini 2.5 Pro & \rankcell{2} & 33.72 & 40.65 & 44.66 & 23.35 & 39.08 \\
Gemini 3.1 Pro & \rankcell{3} & 32.89 & 34.84 & 43.69 & 21.79 & 49.43 \\
Gemini 2.5 Flash & \rankcell{4} & 31.06 & 31.61 & 41.75 & 23.35 & 40.23 \\
Gemini 2.5 Pro + GenS & \rankcell{5} & 25.58 & 33.55 & 32.04 & 13.23 & 40.23 \\
\sectiondivider{Open Models}
Video-LLaVA-7B & \rankcell{26} & 23.75 & 21.29 & 35.92 & 10.89 & 51.72 \\
InternVL3-14B & \rankcell{5} & 37.71 & 42.58 & 21.36 & 37.74 & 48.28 \\
InternVL3-38B & \rankcell{12} & 36.05 & 42.58 & 37.86 & 25.68 & 52.87 \\
InternVL3-78B & \rankcell{1} & \bestopen{\best{41.03}} & \best{\bestopen{43.87}} & 39.81 & 42.02 & 34.48 \\
Qwen2.5-VL-7B & \rankcell{22} & 30.23 & 27.10 & 18.45 & 33.07 & 41.38 \\
Qwen2.5-VL-32B & \rankcell{13} & 35.88 & 34.84 & 29.13 & 33.07 & 54.02 \\
Qwen2.5-VL-72B & \rankcell{2} & 40.37 & 41.29 & 30.10 & 45.14 & 36.78 \\
Qwen3-VL-4B & \rankcell{11} & 36.54 & 34.19 & 33.01 & 32.68 & 56.32 \\
Qwen3-VL-4B-Think & \rankcell{9} & 37.21 & 31.61 & 25.24 & 37.74 & \bestopen{\best{59.77}} \\
Qwen3-VL-8B & \rankcell{15} & 33.72 & 36.13 & 30.10 & 33.85 & 33.33 \\
Qwen3-VL-8B-Think & \rankcell{17} & 31.73 & 34.19 & 33.01 & 25.29 & 44.83 \\
Qwen3-VL-32B & \rankcell{6} & 37.71 & 38.71 & \bestopen{46.60} & 31.91 & 42.53 \\
Qwen3-VL-32B-Think & \rankcell{3} & 40.03 & 38.71 & 22.33 & \bestopen{\best{45.53}} & 47.13 \\
Qwen3-VL-30B-A3B & \rankcell{10} & 36.71 & 30.32 & 22.33 & 42.02 & 49.43 \\
Qwen3-VL-235B-A22B & \rankcell{8} & 37.21 & 37.42 & 25.24 & 39.69 & 43.68 \\
LLaVA-NeXT-Vid-7B & \rankcell{25} & 25.08 & 33.55 & 24.27 & 16.73 & 35.63 \\
LLaVA-NeXT-Vid-34B & \rankcell{21} & 30.40 & 30.32 & 24.27 & 32.68 & 31.03 \\
LlaVA-OneVision-7B & \rankcell{16} & 32.89 & 26.45 & 30.10 & 34.24 & 43.68 \\
LlaVA-OneVision-72B & \rankcell{4} & 38.37 & 35.48 & 25.24 & 38.91 & 57.47 \\
LLaVA-Video-7B & \rankcell{19} & 30.73 & 30.97 & 24.27 & 25.68 & 52.87 \\
LLaVA-Video-72B & \rankcell{7} & 37.54 & 36.77 & 27.18 & 35.80 & 56.32 \\
Perception-LM-1B & \rankcell{24} & 27.74 & 28.39 & 26.21 & 25.29 & 35.63 \\
Perception-LM-3B & \rankcell{18} & 31.40 & 28.39 & 32.04 & 29.96 & 40.23 \\
Perception-LM-8B & \rankcell{14} & 35.38 & 26.45 & 26.21 & 44.75 & 34.48 \\
VideoRefer & \rankcell{23} & 28.57 & 32.90 & 30.10 & 17.51 & 51.72 \\
ArrowRL-7B & \rankcell{20} & 30.56 & 30.97 & 24.27 & 29.18 & 41.38 \\
\bottomrule
\end{tabular}}
\label{tab:maineval}
\vspace{-1.5em}
\end{table}

\Cref{tab:maineval} shows our results. We show the score for the best-performing setting across all video types and visual prompt formats for each model (See App.~\ref{app:results} for more results). 

\medskip \noindent \textbf{Human Performance.}
As shown in Table~\ref{tab:maineval}, humans achieve very high accuracy ($>90\%$ on all question categories), showing that spatio-temporal understanding is essentially second-nature to humans. 
80\% questions received unanimous responses, suggesting that our questions are clear and consistently understood (task-wise breakdowns in App.~\ref{app:results}).

\medskip \noindent \textbf{Proprietary Models.} Unlike humans, both GPT-5~\cite{openai2025gpt5} and Gemini 2.5/3.1 Pro~\cite{google2025gemini25pro} struggled on this task (37.71\% and 33.72/32.89\% respectively): only a modest improvement over the chance baseline (26.74\%), and way below human levels.
Choosing question-relevant frames using GenS~\cite{yao-etal-2025-generative} did not improve Gemini 2.5 Pro's performance.
Thus, unlike these models' success on previous video benchmarks,  proprietary LVLMs struggled with the spatio-temporal understanding required to solve \flatpack.

\medskip \noindent \textbf{Open Models.} 
Among open models, InternVL3~\cite{zhu2025internvl3exploringadvancedtraining} (41.03\%; 95\% bootstrap CI [36.21, 45.64]) performed the best.
Several open models, particularly the Qwen~\cite{bai2025qwen25vltechnicalreport,qwen3vl} and InternVL3~\cite{zhu2025internvl3exploringadvancedtraining} families, are competitive with proprietary models, but performance across open models is highly variable, with some only slightly above the chance baselines.

\medskip \noindent \textbf{Specialized Models.} 
Despite being trained specifically for fine-grained region-specific questions, PerceptionLM~\cite{cho2025PerceptionLM} and VideoRefer~\cite{yuan_videorefer_2025} perform poorly, likely due to a mismatch between
their training data, which feature relatively simple videos with few fine-grained interactions, and the complex, multi-part interactions in \flatpack.
Still, PerceptionLM-8B is competitive with much larger models (e.g., Qwen2.5-VL-32B), suggesting value in training on data rich in fine-grained interactions.
Temporal sensitivity also helps ArrowRL~\cite{xue2025seeing} beat its base checkpoint (Qwen2.5-VL-7B), especially on temporal ordering (\textsc{TOrd}).
We also analyze correlations between model performance and factors such as video duration and difficulty (See App.~\ref{app:results}).

\medskip
\noindent
Taken together, these results indicate that current LVLMs--proprietary, open, and specialized--remain far from achieving the strong, reliable spatio-temporal understanding skills that humans demonstrate on \flatpack.

\section{Analysis}
\label{sec:analysis}
We now examine why LVLMs struggle on our benchmark. 
Unless stated otherwise, we use a standard open LVLM, Qwen2.5-VL-72B for analysis (See App.~\ref{app:analysis} for experimental details).
Other open LVLMs, like InternVL3-78B, yield similar overall takeaways (App.~\ref{app:internvlanalysis}).

\subsection{Linguistic Prompt Engineering}
\label{sec:promptengg}

Previous work has shown that linguistic prompting strategies can improve model performance on language-based reasoning tasks~\cite{wang2023demo2code,surismenon2023vipergpt}. 
We evaluate whether such techniques also offer any benefit for the spatio-temporal reasoning required in our benchmark. 
We use key-frame videos for this experiment with Mixed-media prompts (as video type had little impact on accuracy, key-frame videos were faster to evaluate, and Mixed-media was the best prompt type for this LVLM; See \cref{sec:visualprompt}).
We consider the following approaches:
\begin{itemize}
    \item \textbf{Zero-shot Chain-of-Thought Prompting}~\cite{zeroshotcot} (\textsc{ZS-CoT}): We prompt the LVLM to generate an explanation of their answer choice. 
    We modify the task instructions to include ``\textit{Please explain your answer step-by-step}''.
    This baseline also uses greedy decoding.
    \item \textbf{Self-consistency w/. CoT}~\cite{wang2023selfconsistency} (\textsc{SC-CoT}): We extend \textsc{ZS-CoT} with temperature sampling, generating 5 responses and then select the final answer with majority voting.
\end{itemize}
\begin{table}[h]
\vspace{-0.5em}
\centering
\small{
\caption{\textbf{Effect of Lingustic Prompting Strategies.} Both ZS-CoT and SC-CoT fail to improve performance on \flatpack.}
\renewcommand{\arraystretch}{1.2}
\renewcommand{\tabcolsep}{1.2mm}
\resizebox{\linewidth}{!}{
\begin{tabular}{lccccc}
\toprule
 & \textbf{Micro Avg.} & \textbf{\textsc{TOrd}} & \textbf{\textsc{TLoc}} & \textbf{\textsc{Track}} & \textbf{\textsc{Mate}} \\
\midrule
-- & 40.19	& 40.64 & 30.09	& 45.52	& 35.63 \\ %
ZS-CoT & 39.20 & 38.06 & 30.10 & 43.19 & 40.23 \\
SC-CoT & 32.23 & 32.90 & 24.27 & 35.02 & 32.18 \\
\bottomrule
\end{tabular}}
\label{tab:cot_comparison}}
\vspace{-1em}
\end{table}

\Cref{tab:cot_comparison} shows their performance, comparing them with vanilla prompts used in \cref{sec:eval}.
Both approaches hurt performance, suggesting that unlike language-based reasoning tasks, CoT-based prompting
does not improve spatio-temporal visual understanding required for our benchmark.

\subsection{Effect of Visual Data Processing}
\label{sec:visualprompt}
\begin{figure*}
    \centering
    \includegraphics[width=0.95\linewidth]{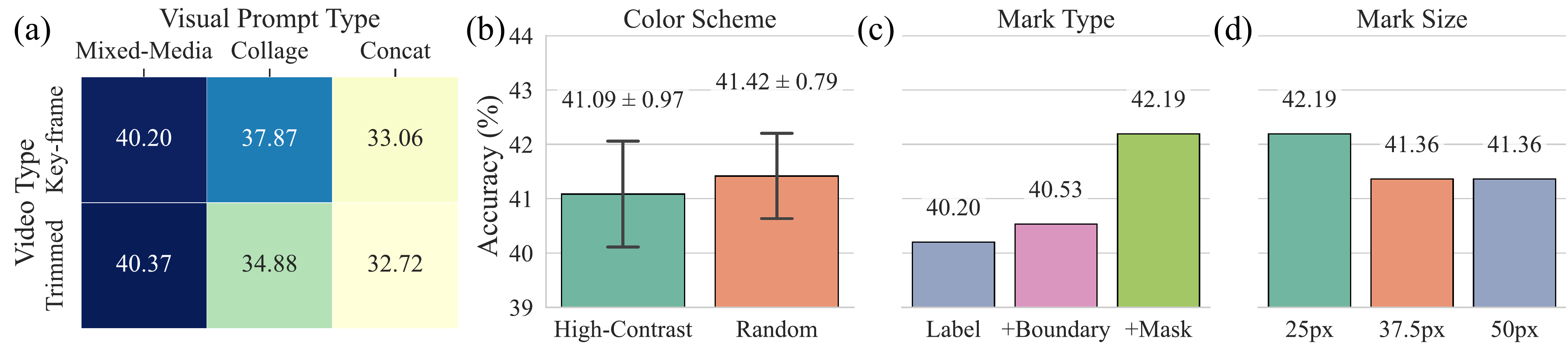}
    \caption{\textbf{Visual Data Ablation.} We study the effect of different strategies of providing the visual prompt and video processing (a). Next, we analyze how the (b) color scheme, (c) mark type, and (d) mark size affect the LVLM's performance on our benchmark.}
    \label{fig:visualpromptablation}
    \vspace{-1.5em}
\end{figure*}

Past work~\cite{Liu_2025_CVPR} has also shown that LVLMs are sensitive to visual data processing.
We therefore evaluate the impact of the various choices around the visual input.

First, we look at the impact of trimmed vs. key-frame videos, and mixed-media vs. collage and concat prompts (\cref{fig:visualpromptablation}(a)).
We find that the video type does not have a big impact.
However, among visual prompt types, mixed-media approaches perform the best, likely because of the mixed image and video training data for Qwen2.5-VL~\cite{bai2025qwen25vltechnicalreport}.

Next, we look at visual prompt rendering parameters.
We use key-frame videos since the video type has little impact, and they are faster to evaluate.
To create our visual prompts, we greedily sampled each mask color to be maximally distinct from the closest color already chosen and its underlying pixel to ensure visual distinctiveness.
We also added a 2-pixel boundary around each mask to increase its saliency.
We ablate through three axes:
\begin{itemize}
    \item \textit{\textbf{Color Scheme}}: Does greedily selecting a high-contrast color perform better than randomly picking a color? We render visual prompts across 3 runs and compare. 
    \item \textit{\textbf{Mark Type}}: We first overlay only the part IDs on the image, then the mask outline and finally, the mask itself.
    \item \textit{\textbf{Mark Size}}: We gradually increase the marks' font size.
\end{itemize}

We find that high-contrast colors (\cref{fig:visualpromptablation}(b)) and mark size (\cref{fig:visualpromptablation}(d)) have limited impact, but it is important to render all three: the label, the boundary and the mask (\cref{fig:visualpromptablation}(c)).

\subsection{Do LVLMs utilize temporal context effectively?}
\label{sec:imageonly}

\begin{table}[t]
\centering
\caption{\textbf{Image-only Prompt.} Performance of the LVLM using image-only prompts, along with the change ($\Delta$) in performance from when the video is included in the prompt.}
\setlength{\tabcolsep}{5pt}
\renewcommand{\arraystretch}{1.18}
\renewcommand{\tabcolsep}{1.2mm}
\resizebox{\linewidth}{!}{
\begin{tabular}{lccccc}
\toprule
 & Micro Avg. & \textsc{TOrd} & \textsc{TLoc} & \textsc{Track} & \textsc{Mate} \\
\midrule
\textit{Video+Image}
& 40.20	& 40.65 & 30.10 & 45.53 & 35.63 \\
\hdashline
Image-only
& 31.40 & 40.65 & 34.95 & 21.01 & 41.38 \\
\quad$\Delta$
& \dneg{--8.80} & 0.00 & \dpos{+4.85} & \dneg{--24.51} & \dpos{+5.75} \\
\hdashline
Shuffled Parts
& 29.96 \scriptsize{$\pm$ 0.69} & 35.70 \scriptsize{$\pm$ 3.55} & 36.25 \scriptsize{$\pm$ 4.04} & 20.62 \scriptsize{$\pm$ 1.68} & 39.85 \scriptsize{$\pm$ 2.65} \\
\quad$\Delta$
& \dneg{--10.24} 
& \dneg{--4.95}
& \dpos{+6.15}
& \dneg{--24.90}
& \dpos{+4.21} \\
\hdashline
\textbf{\textit{Human Perf.}} & 42.69 & 30.32 & 45.63 & 48.24 & 44.82 \\
\bottomrule
\end{tabular}}
\label{tab:imageonly}
\vspace{-1.5em}
\end{table}

The high human performance on our benchmark shows that the questions can be easily solved using the provided information, namely the assembly video and the visual prompt.
Yet, most of the SOTA LVLMs struggle on our benchmark, particularly when the task requires careful temporal understanding, such as \textsc{Track}. 
The video is the main source of temporal information about the assembly in the question.

This leads us to ask the question: \textit{Are LVLMs using the videos effectively?}
To answer this question, we evaluated these models on an image-only version of our benchmark, i.e., the model was only provided the question text and the visual prompt image(s). 
We also measured human performance on this version (similar to \cref{sec:eval}).
\Cref{tab:imageonly} shows the performance of the LVLM on this image-only version, and the change in their performance from the full evaluation, along with human performance.
The sharp drop in human performance ($>$50\%) shows that the questions do require videos to answer.
We also observe that the overall performance of the model drops severely (8.80\%), but mostly due the \textsc{Track} sub-task. 
Accuracy on other tasks stays the same or improves, indicating that the LVLM \emph{does not use the video effectively}, while humans use the video to answer.

Task-wise, performance improves on the \textsc{Mate} and \textsc{TLoc} sub-tasks.
This suggests that the model used its superior image-understanding ability to recognize the part types (e.g., legs, backrest, etc.) and their positioning (e.g., hands holding/close to the next/previous part connected) in the prompt image, along with commonsense reasoning.
Higher human performance on these sub-tasks, compared to \textsc{TOrd} which requires long temporal context, also suggests that such image- and commonsense-based shortcuts might be feasible.

\paragraph{Part ID Bias} 
We also observe no change in performance on \textsc{TOrd} instead of an expected decline.
We hypothesized that the counterintuitive observation was due to a bias in the questions towards the integer values of part IDs corresponding to the correct answer of \textsc{TOrd} questions.
To verify this, we try 3 separate runs where we shuffled the part IDs for each part with different seeds, and report the average performance.
As shown in \cref{tab:imageonly}, shuffling the part IDs led to a decline in performance on \textsc{TOrd}, suggesting that the order of Part IDs was one of the short-cuts that the LVLM was exploiting.

\medskip
\noindent
Overall, these results suggest that LVLMs are not using the videos effectively.

\subsection{Probing Errors with Self-Explanations}
\label{sec:selfexpl}
\begin{figure}
    \centering
    \includegraphics[width=\linewidth]{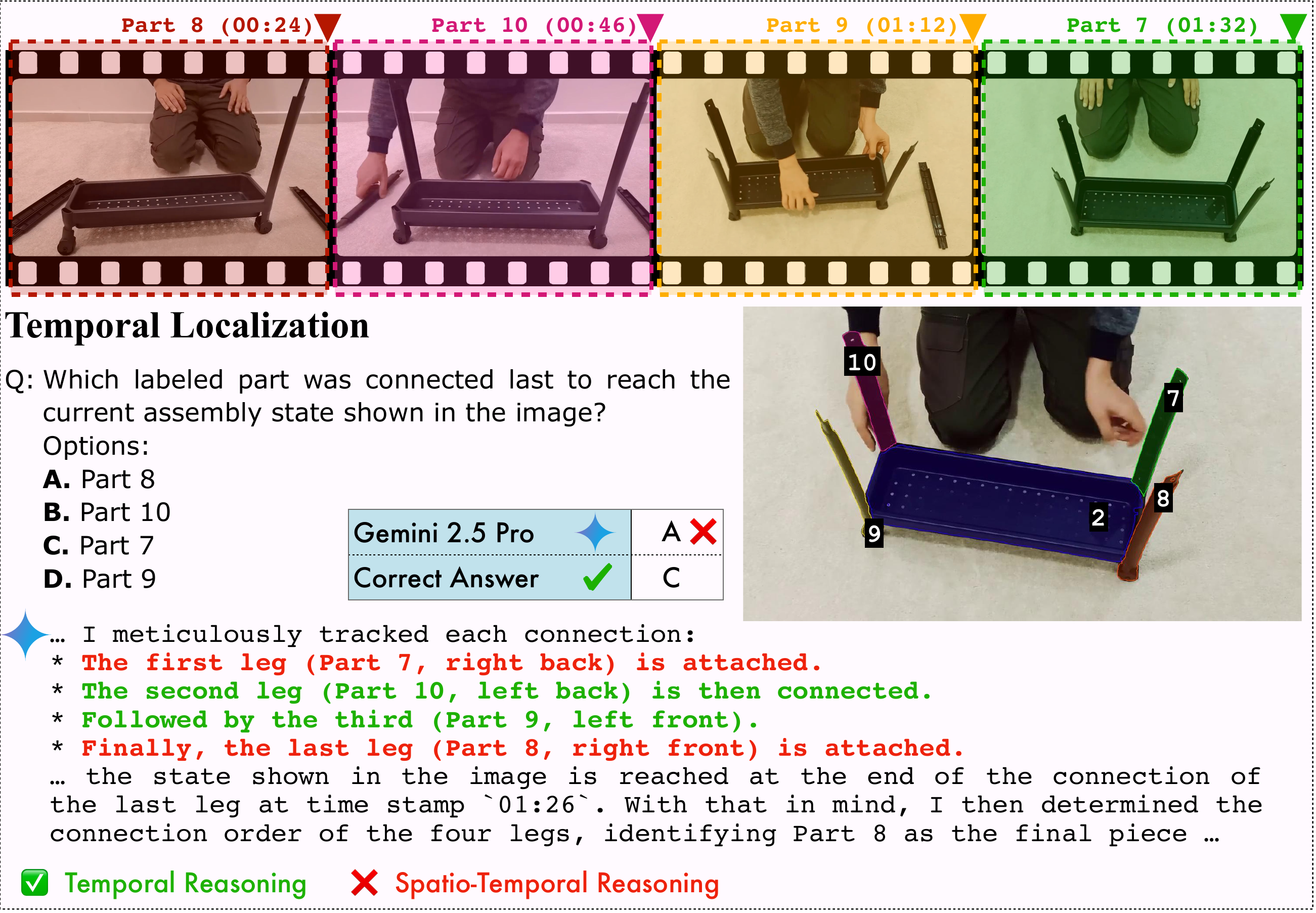}
    \caption{\textbf{Self-probing Explanations.} Qualitative example from Gemini 2.5 Pro. We highlight the video with the relevant connection events for clarity. We can observe that the model looks at the video, but makes an error due to gaps in its spatio-temporal reasoning.}
    \label{fig:probeexpl}
    \vspace{-1.3em}
\end{figure}

Since LVLMs are failing to effectively utilize the spatio-temporal context in videos, we perform a deeper investigation into such mistakes.
To investigate this, we examined the explanations produced by the models while solving our tasks.
However, in preliminary experiments, we found that the rationales generated by the open LVLM were largely paraphrases of its chosen answer, offering little insight into its reasoning.
By contrast, the internal thought summaries~\cite{google2025geminiThinking} produced by Gemini 2.5 Pro were substantially richer--featuring time-stamped reasoning and explicit mentions of connection events--which made it far easier to pinpoint the causes of its errors. We therefore base our analysis on these summaries.

\paragraph{Qualitative Example}
Figure~\ref{fig:probeexpl} shows a shortened self-explanation from Gemini 2.5 Pro.
Human annotators answered this question correctly.
The model demonstrates some temporal localization ability, correctly identifying the timestamp of the final connection event but fails to constrain its subsequent reasoning to that point.
It also infers the order of Parts 10 and 9 correctly.
However, it mis-tracks the first part to be connected--predicting it as the right back leg (Part 7) instead of the right front leg (Part 8)--inter-changing their position, leading to a wrong answer. For more examples see our project site: \projectsite[self-explanations].

\paragraph{Error Types from Rationales} 
In order to do a quantitative analysis on the error types based on the self-explanations, we selected 200 questions where the model got the answer wrong, sampled equally from all 4 question categories.
We asked human annotators to write 1-2 sentences on why they think the model committed an error based on the thought summary.
Then, for all the questions we collated the question text, the model's thoughts, the annotators' comments, asked an LLM (Gemini 2.5 Pro) to identify 5 error categories that cover the annotations, and assign each question to one or more categories.
In this way, we arrived at the following categories ($\%$ reflects share within the selected questions):
\begin{itemize}
    \item \textbf{Object Grounding ($37.28\%$)}: Failure to correctly identify an object across the image and video.
    \item \textbf{Spatio-Temporal Reasoning ($32.45\%$)}: Error in tracking an object's identity through spatial transformations like camera movement, object rotation, or scene cuts. 
    \item \textbf{Temporal Reasoning ($17.98\%$)}: The model gets the chronological sequence of object interactions wrong.
    \item \textbf{Physical Interaction ($7.89\%$)}: The model misjudges contact, support, or other physical interactions between parts.

    \item \textbf{Language \& Logic ($4.38\%$)}: 
     Misinterpreting instructions or drawing faulty conclusions from correct observations.
\end{itemize}

Object grounding and spatio-temporal reasoning are the major error sources,
 suggesting that the model struggles to understand the fine-grained regional references from the visual prompt and  tracking the references through the video.

\subsection{Can Task Decomposition Improve Spatio-Temporal Reasoning?}

Results so far show that our task is very difficult for current LVLMs to solve in a zero-shot manner.
However, we find that most of the tasks in our benchmark can be solved by the use of two primitives: tracking objects (\textsc{Track}) and contact reasoning (whether two parts are connected in a particular frame). 
For instance, consider a Temporal Ordering (\textsc{TOrd}) question shown in~\cref{fig:snapshot}, where we want to determine the sequence in which parts are connected during assembly. 
Suppose we have an oracle that provides accurate segmentation maps for each relevant part at every timestamp in the assembly video (the tracking problem), along with their connection status (the contact reasoning problem). 
We can then iterate through all frames, checking the presence and connection status of each part from the visual prompt. 
By recording the timestamps at which each part becomes connected and sorting them chronologically, we can effectively answer this question. 
This raises an important question: Can we design vision systems that can solve the tasks in our benchmark by decomposing them into tracking and contact-reasoning?

\subsubsection*{\textbf{\tva}: An Agentic Baseline}
\label{sec:tva}

\begin{figure*}
    \centering
    \includegraphics[width=\linewidth]{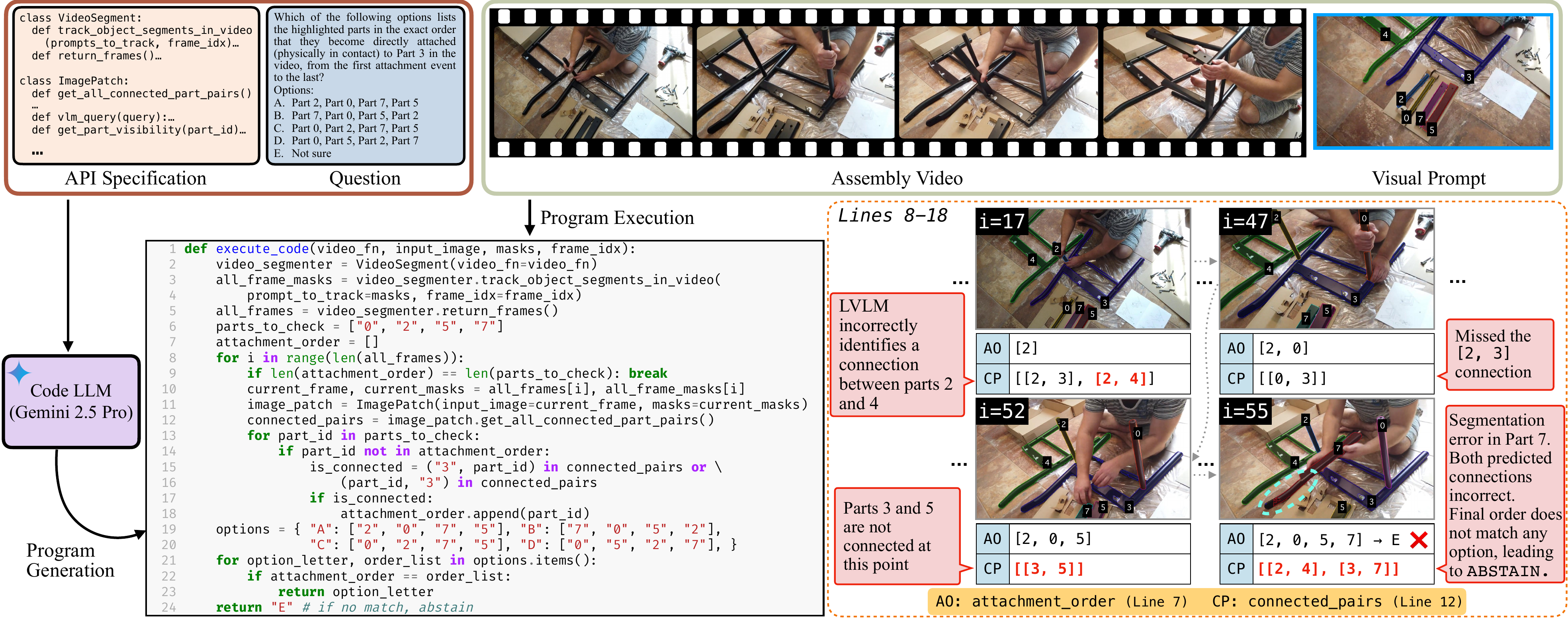}
    \caption{\textbf{Temporal Video Agent.} An overview of our agentic baseline. First, a Code LLM uses the API specification and the input question to generate a program. The generated program uses the assembly video and the visual prompt's frame index and mask to produce a response for the question. We also show an example trace for a question. We can analyse the execution trace to pin-point the sources of error.}
    \label{fig:tva}
    \vspace{-1.5em}
\end{figure*}

To test this, we propose a baseline, \textbf{Temporal Video Agent (\tva)}, a visual programming agent~\cite{surismenon2023vipergpt} that has access to tracking and contact-reasoning tools for solving assembly-based tasks as proposed in our benchmark. 
Following ViperGPT~\cite{surismenon2023vipergpt},  we provide a pythonic API to a \textit{Code LLM} (Gemini 2.5 Pro~\cite{google2025gemini25pro}), that also receives the question as input and generates a program that makes use of the provided API to write a function that can potentially produce an answer.
The API includes a video object segmentation function built on top of SAM2~\cite{ravi2025sam}, and a VLM-query function (supported by Qwen2.5-VL-32B) that can ask a question related to an image (for resolving contact-reasoning queries) among other tools.
While all our questions have a valid answer, it is possible for \tva{} to end up with a valid program that produces an answer outside of the given options due to tool failures.
Hence, we add an \textsc{abstain} option (``\textit{Not Sure}'') to each question.
We also record the success rates of program execution for the code generated by the orchestrator LLM (See~\cref{tab:agent_metrics}).

\begin{wraptable}{r}{0.23\textwidth}
\centering
\footnotesize
\vspace{-1em}
\caption{\textbf{Agent performance.} \textit{Acc. (Answered)} is accuracy over non-abstained questions}
\begin{tabular}
{@{}l S[table-format=2.2]@{}}
\toprule
\textbf{Metric} & \textbf{Value (\%)} \\
\midrule
Accuracy             & 11.79 \\
Abstained            & 62.29 \\
Acc.\ (Answered)  & 31.27 \\
Program failed       &  3.32 \\
\bottomrule\\
\end{tabular}
\label{tab:agent_metrics}
\vspace{-3em}
\end{wraptable}

Overall \tva{} performs quite poorly. 
However, we did find that it is able to correctly answer 11.48\% of the questions missed by the LVLM.
Figure \ref{fig:tva} overviews the method and shows a qualitative example of a generated program and its execution\footnote{More examples at \projectsite[tva/]}.
We observed that inaccuracies in the available tools were the primary causes of errors.
We further analysed the performance of individual tools to concretely study the extent of the problem.

\paragraph{Contact-Reasoning Issues} 
For each pair of visible parts in the visual prompt, we prompted the LVLMs with a binary \texttt{Yes/No} question that asks the models if the two parts are connected as they would be in the final assembly.
The ground-truth labels for these questions are mined from the key-frame annotations in \imaw{}~\cite{liu2024ikea}.
In this manner, we obtain 1500 questions (750 for each \texttt{Yes} and \texttt{No}). We used Qwen2.5-VL-32B for this evaluation as that was LVLM used in~\tva{}.
Our results confirmed what the~\tva{} results hinted: while the LVLM achieves 64.33\% accuracy on this task, the accuracy on the \texttt{Yes} questions is only \textbf{\textcolor{red}{52.93\%}}, only slightly better than random chance (For more details, see App.~\ref{app:tvaresults}).  

\paragraph{Tracking Issues}
We use the annotated ground-truth segmentations for selected frames in the videos to compute the accuracy of SAM2. We prompt  SAM2~\cite{ravi2025sam} with our manually curated segmentation masks in one frame to track them through the video, evaluating these tracks in a second annotated frame. 
SAM2 achieves a fairly low average IoU of \textbf{\textcolor{red}{0.28}} (see App.~\ref{app:tvaresults} for details). 
This suggests that tracking objects in the wild is also a bottleneck for \tva{}~and by extension, would also prove a challenge for LVLMs that must do such tracking implicitly to understand the videos.

\medskip
\noindent
These results suggest that beyond the benchmark questions, even simpler tasks on the furniture assembly domain are a challenge for current state-of-the-art.

\section{Conclusion}

We proposed \flatpack{}, a VidQA benchmark that tests LVLMs' performance on spatio-temporal understanding.
Our analysis reveals key bottlenecks of LVLMs in fine-grained video understanding particularly spatio-temporal reasoning (e.g. keeping track of parts through occlusions, scene cuts, etc.) and region-specific grounding.
We also study whether an agentic decomposition of the task can help, but find that current vision tools share similar limitations: even specialized tracking models struggle, and LVLMs fail on simpler subproblems such as contact reasoning.
Future avenues of work include exploring task-specific fine-tuning on synthetic simulated data, improving visual prompting techniques for regional understanding, and more sophisticated agentic pipelines that  can leverage low-level signals like 3D geometry, depth, etc. for improved performance.

\section*{Acknowledgments} 

We thank Brihi Joshi, Alice Lu, Gemmechu Hassena, Shamus Li, Snehal Bhagat, Katie Luo, Xinrui Liu, Sushrut Sudarshan Surve, Rishabh Madan, Abhishek Vijaya Kumar, Chuanruo Ning, Samuel Speas, Ruyu Yan, Kuan Wei Huang, and Rajeev Datta for their contributions as human annotators and evaluators.
We thank Yunong Liu for assistance with the~\imaw{} dataset.
We also thank Wei-Chiu Ma, Yihong Sun, Benlin Liu, Jieyu Zhang, Zixian Ma, Divy Thakkar, Raushan Turganbay, Aritra Roy Gosthipaty, and members of the Cornell Graphics and Vision group for insightful discussions, feedback, and support throughout this project.
This work was funded in part by the National Science Foundation (IIS-2144117, IIS-2403015, IIS-2211259).
We also acknowledge support from the Google Gemini Academic Program.

{
    \small
    \bibliographystyle{ieeenat_fullname}
    \bibliography{main,fpb}
}

\appendix
\setcounter{figure}{0}
\renewcommand{\thefigure}{S\arabic{figure}}
\setcounter{table}{0}
\renewcommand{\thetable}{S\arabic{table}}
\clearpage

\setcounter{page}{1}
\maketitlesupplementary

\section{Benchmark Details}
\label{app:benchdetails}

\subsection{Additional Benchmark Statistics}
\label{app:qualexamples}
 
The dataset contains videos of 24 unique furniture items, consisting of 3-19 parts (average of 7 parts).
Keyframe videos are pre-extracted at 1 FPS, whereas trimmed videos retain their original variable frame rates, averaging 28.98 FPS. 
The average duration of keyframe videos is about 6 minutes, compared with 2.98 minutes for trimmed videos. 
Based on keyframe videos and connection-event annotations derived from \imaw{}\cite{liu2024ikea}, the minimum number of frames required to answer a question is 113.7 on average (median: 69.5). 
Similarly, for \track~questions, the average interval between the visual prompts is 141.1 frames (median: 76).

\subsection{Incomplete Segmentations in \imaw{}}
\label{app:vp}

\begin{figure}
    \centering
    \includegraphics[width=\linewidth]{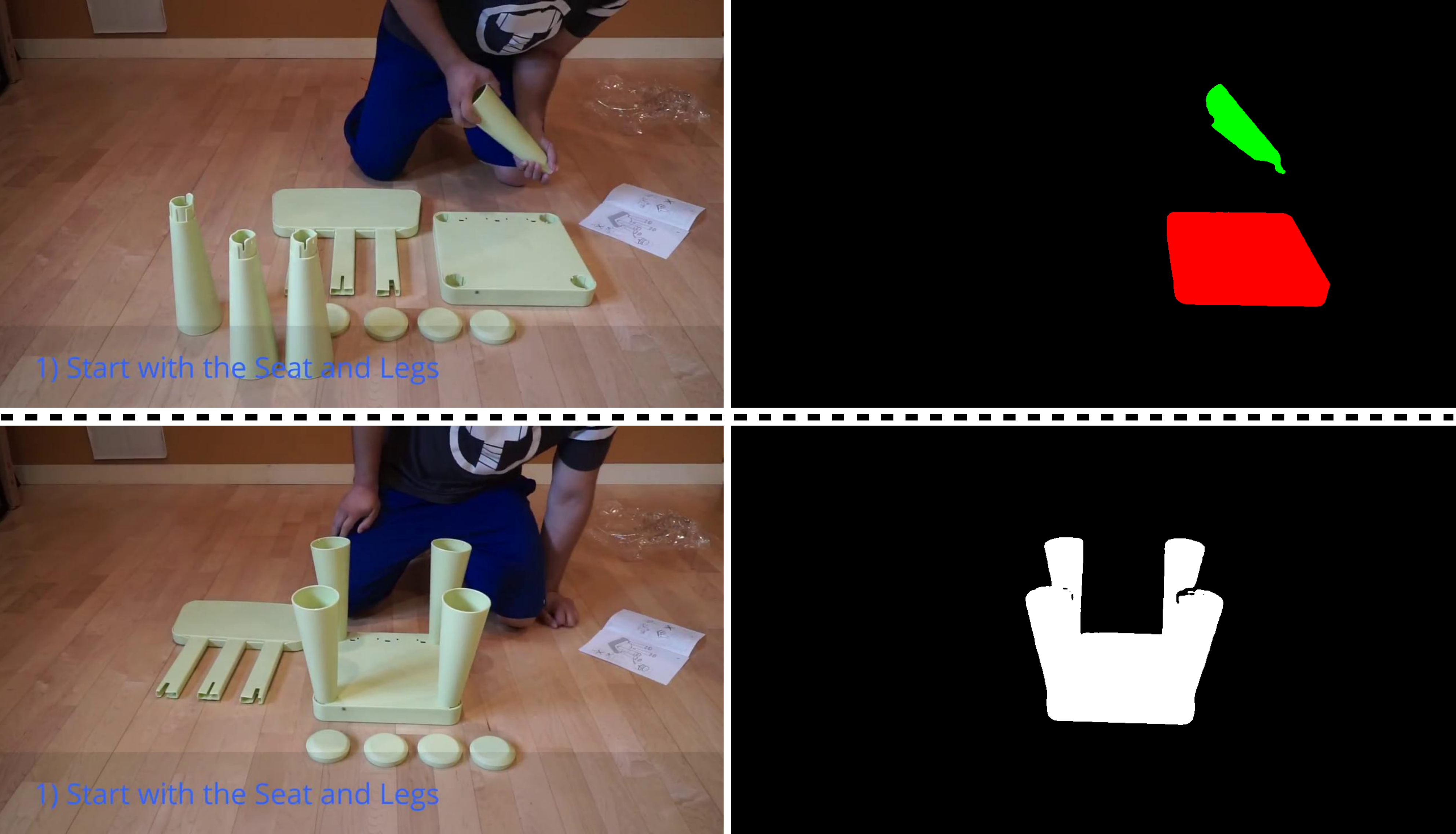}
    \caption{\textbf{Incomplete Segmentations in \imaw{}~\cite{liu2024ikea}.} (Top) Segmentations are only provided for the parts that are about to be connected, i.e., the seat panel and the leg. Parts not being interacted with (the other legs and the backrest) are not annotated. 
    (Bottom) Furthermore, instead of granular segmentations of each individual part, \imaw{} only has masks at the sub-assembly level.}
    \label{fig:incompletesegs}
    \vspace{-1.5em}
\end{figure}

In~\cref{sec:dataset}, we mentioned that~\imaw{}~\cite{liu2024ikea} has incomplete segmentation annotations. Here, we elaborate this statement:
\begin{enumerate}
    \item  We found that segmentation annotations are only provided for parts that are in the process of being connected in a particular key frame. 
    This limits the questions to these particular key frames, and even so, to only the parts annotated therein.
    \item Furthermore, \imaw{} only includes information at the sub-assembly granularity, which precludes questions that one might want to ask about specific parts.
\end{enumerate}

Figure~\ref{fig:incompletesegs} shows some examples of available segmentations in \imaw{}~to illustrate these issues. Thus, we annotate our own segmentation maps as described in~\cref{sec:dataset}.

\subsection{Manual Question Curation}
\label{app:questioncuration}
\begin{figure}
    \centering
    \includegraphics[width=\linewidth]{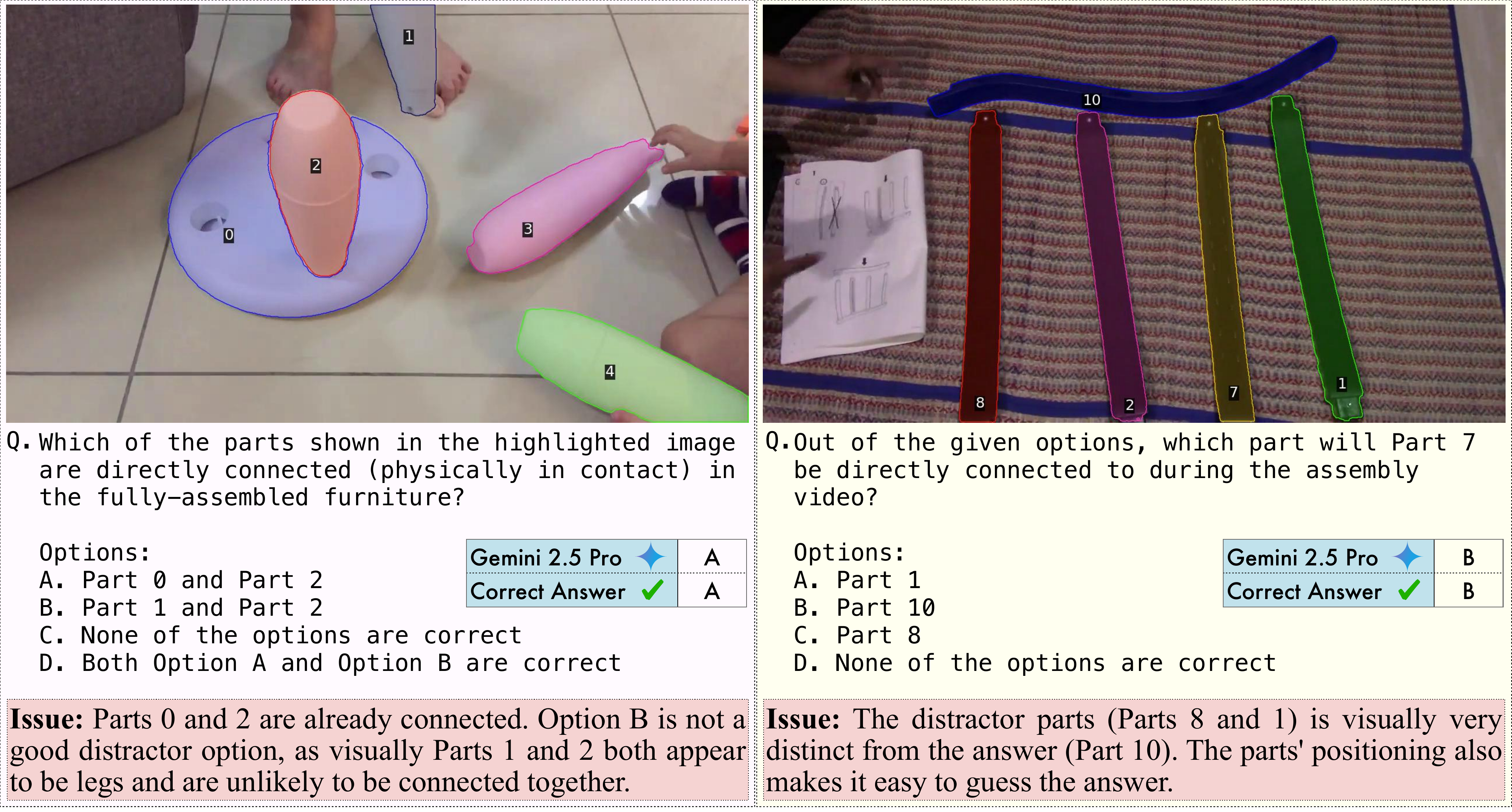}
    \caption{\textbf{Issues with Auto-generated Questions.} We show some examples to highlight the issue of auto-generated questions where the answer can be arrived at using the visual prompt alone and commonsense reasoning. Observe that we can arrive at the correct answer without requiring any temporal cues from the video.}
    \label{fig:autogenquestions}
    \vspace{-1em}
\end{figure}

\paragraph{Pitfalls of Auto-generated Questions}~\Cref{fig:autogenquestions} shows some examples of auto-generated questions, where the question can be solved using the visual prompt alone. 
In the first example (left), Part 0 is already connected to Part 2 in the prompt. 
Furthermore, Parts 1 and 2 are both clearly legs and unlikely to be connected given the structure of Part 0.
Overall, this frame is unsuitable for a \mating{} question.
Similarly, in the second example, the distractor parts (Parts 1 and 8) are visually quite distinct from Part 10.
Their positioning, as if they are about to be inserted into Part 10 (not Part 7), also makes it easy to answer this question without any temporal context.
Observe that even without the videos, the correct answer is easy to infer from the visual prompt and commonsense reasoning alone.
Such examples motivated us to manually curate our questions.

\paragraph{Question Templates} 
\Cref{tab:assembly_tasks} shows the full list of question templates used by our annotators.
Annotators had to populate the parts highlighted in red with part IDs and provide options that they believe are challenging (parts with similar appearance, parts that are connected much later in the assembly compared to the state shown in the visual prompt, etc.) 
All templates require up to 4 options, except for the second template in \mating{}, which is a binary question.

\begin{table*}[t]
\small
    \centering
    \begin{tabular}{r|p{15cm}}
        \textbf{Task} & \textbf{Question Template} \\
        \hline
        \mating & 
        \begin{itemize}
        \vspace{-0.3cm}
            \item \textit{Out of the given options, which part will \textcolor{red}{\{query\_part\}} be directly connected to during the assembly video?}
            \item \textit{Are \textcolor{red}{\{query\_part1\}} and \textcolor{red}{\{query\_part2\}} connected in the fully-assembled furniture?}
            \item \textit{Which of the parts shown in the highlighted image are directly connected (physically in contact) in the fully-assembled furniture?}
        \end{itemize} \\
        \hline
        \localization & 
        \vspace{-0.3cm}
        \begin{itemize}
            \item \textit{Which of the highlighted parts are connected (brought in direct physical contact) to \textcolor{red}{\{query\_part\}} next starting from the assembly state shown in the image?}
            \item \textit{Which highlighted part is the last to connect (brought in direct physical contact) to \textcolor{red}{\{query\_part\}} in the video?}
            \item \textit{Which highlighted pair of parts is connected first in the video?}
            \item \textit{Which highlighted pair of parts is connected last in the video?}
            \item \textit{Which labeled part was connected last to reach the current assembly state shown in the image?}
            \item \textit{Which labeled part will be connected next to the current assembly state shown in the image?}
        \end{itemize} \\
        \hline
        \ordering & 
         \vspace{-0.3cm}
        \begin{itemize}
            \item \textit{Which of the following options lists the highlighted parts in the exact order that they become directly attached (physically in contact) to \textcolor{red}{\{query\_part\}} in the video, from the first attachment event to the last?}
            \item \textit{Each option below lists a sequence of part connections. Which of these options gets the temporal ordering of the connections correct?}
        \end{itemize} \\
        \hline
        \track & 
         \vspace{-0.2cm}
        \begin{itemize}
            \item \textit{Identify the correct set of matches from Image A to Image B.}
            \item \textit{Identify which part in Image B matches Part \textcolor{red}{\{query\_part\}} in Image A.}
            \vspace{-0.2cm}
        \end{itemize} \\
    \end{tabular}
    \caption{\textbf{Question Templates.} We replace the \textcolor{red}{highlighted} part in the question template from scene to scene to construct our benchmark. 
    }
    \vspace{-.5cm}
    \label{tab:assembly_tasks}
\end{table*}

\section{Evaluation Details}
\label{app:evaldetails}

Here we provide additional experimental details for~\cref{sec:eval}, including video processing and visual prompt construction.

\paragraph{Video Data Processing} All videos are resized to a height of 480 px and a width of 640 px. 
We dump all videos at 1 FPS before passing them as input to the model.

\paragraph{Video Subsampling} 
During evaluation, we first decode each video at its stored frame rate and expose all frames in that stored video to the model pipeline, so that no connection event is discarded a priori. 
Thus, keyframe videos provide all 1 FPS keyframes, while trimmed videos provide all frames at their original variable frame rates. 
Any subsequent frame selection is then performed by the individual model pipeline, subject to its context-length constraints, using a common 1 FPS temporal view for both video types.
The videos in our benchmark are quite long (average duration of 6 minutes for key-frame videos at 1 FPS, i.e., 360 frames on average). 
Typically, it is not possible to pass all the frames as input to models due to their limited context length.
Hence, we uniformly subsample the frames before passing them as input.
Subsampling frames uniformly from a fixed frame rate of 1 FPS is similar to the FPS-sampling strategy advocated by recent works~\cite{zohar_apollo_2024,liu_oryx_2025}.
For the number of frames to subsample, we followed the official guides/cookbooks released by the models wherever possible.
Below we provide some details about subsampling choices:
\begin{itemize}
    \item \textbf{Proprietary Models}: For Gemini~\cite{geminiteam2025geminifamilyhighlycapable}, we provide the entire video as input. 
    For trimmed videos that exceed the duration limit of 1 hour for Gemini, we dump the videos at the minimum FPS where the duration limit is satisfied. 
    For GPT-5~\cite{openai2025gpt5}, we provide 499 frames (498 in case of \track{} questions), as per the API token limit, which accommodates 500 frames.
    \item \textbf{Open Models}: 
    For the Qwen family of models~\cite{bai2025qwen25vltechnicalreport,qwen3vl} (including ArrowRL~\cite{xue2025seeing}), we sample a maximum of 768 frames, while maintaining a context length of 20480 tokens, as suggested by their official cookbook~\cite{qwen3vl}.
    For all other open models, we subsampled 32 frames, except Video-LLaVA~\cite{lin2023video} (8 frames), as per the setting used within their respective training and evaluation pipelines.
\end{itemize}

Note that, for all models, during subsampling of \textit{Concat} prompts, we modify the subsampling procedure so that the first two frames (containing the visual prompts) are retained, while maintaining the total number of frames.

\paragraph{Visual Prompt Construction} As mentioned in~\cref{sec:eval}, we use three different settings for constructing the visual prompt -- \textit{Mixed-Media}, \textit{Collage}, and \textit{Concat}. 
Examples of the complete visual inputs for different prompt settings can be found on the project site\footnote{\projectsite[visual-prompts/]}

The images that constitute the visual prompts are also resized to a resolution of $480 \times 640$.
For \track{} questions, the jumbled image prompt (Image B in~\cref{fig:snapshot}) is provided as the \textit{second} image in the input for \textit{Mixed-Media}, as the middle image in each frame for \textit{Collage}, and as the second frame of the video for \textit{Concat}. 
These details are also conveyed to the model in the task instructions provided along with the input.

\paragraph{Evaluation Setup for Specialized Models} We set up the evaluation for specialized models as follows:
\begin{itemize}
    \item \textbf{ArrowRL}~\cite{xue2025seeing}: As discussed, for ArrowRL~\cite{xue2025seeing}, we use a similar evaluation setup to Qwen 2.5-VL, as that is the base model from which it was created.
    \item \textbf{PerceptionLM}~\cite{cho2025PerceptionLM}: For PerceptionLM~\cite{cho2025PerceptionLM}, we subsample 32 frames from the video, similar to the evaluation settings described in the paper. 
    We evaluate PerceptionLM only on \textit{Collage} and \textit{Concat} prompts because it lacks support for \textit{Mixed-Media} prompts.
    \item \textbf{VideoRefer}~\cite{yuan_videorefer_2025}: For evaluating VideoRefer~\cite{yuan_videorefer_2025}, instead of using our prompting strategies, we follow its inference recipe by directly giving the part segmentations and the frame indices for the visual prompts as input to the model.
    \item \textbf{GenS}~\cite{yao-etal-2025-generative}: For GenS~\cite{yao-etal-2025-generative}, we found that the original model can only handle up to 256 frames. 
    Hence, to evaluate longer videos, we split the video into chunks of 256 frames and obtained a relevance score for frames in each chunk. 
    We filtered the frames in each chunk using a score threshold of 4 and provided the remaining frames as input to the base model (Gemini 2.5 Pro).
\end{itemize}

\paragraph{Task Instructions} 
In addition to the video, the visual prompts, and the question, the input contains task instructions that describe the input visual data to the model and specify the output format that the model should follow. 
\Cref{fig:taskdesc,fig:taskdesccollage,fig:taskdescconcat} show the task instructions that we use in our evaluations. 
When asking \track{} questions, we make adjustments to the input descriptions, informing the model that there are two visual prompts instead of one.
Note that the task instructions are not a part of the benchmark. 
We welcome users of the benchmark to try out different phrasings of the instructions.
For reproducibility, we release the source code used in our evaluations at: \href{https://github.com/justachetan/flat-pack-bench}{\nolinkurl{github.com/justachetan/flat-pack-bench}}.

\begin{figure*}[t]
\begin{tcolorbox}
[title=Task Instructions for Mixed-Media Prompts]
{\color{myorange}
You are a furniture-assembly expert. You are given:
\begin{enumerate}[nosep,leftmargin=*,label=\arabic*.,ref=\arabic*]

  \item A video of a furniture assembly in progress.

  \item A single labeled frame from the same video, displaying numeric IDs on each visible furniture part. 
  Call this a visual prompt.

  \begin{tcolorbox}[
    colback=gray!10,
    colframe=gray!50,
    boxrule=0.3pt,
    left=1mm,
    right=1mm,
    top=0.5mm,
    bottom=0.5mm,
    title=\footnotesize For \track{} Questions,
    coltitle=black,
    fonttitle=\bfseries,
    coltext=myorange,
    enhanced,
  ]
  \footnotesize
  \begin{enumerate}[nosep,leftmargin=*,label=\arabic*.,ref=\arabic*,start=2]
      \item  Two labeled frames from the same video, called Image A and Image B, each displaying numeric IDs on visible furniture parts. Note: The numeric IDs in Image A and Image B are not necessarily the same; the same ID may refer to different parts in each image.
  \end{enumerate}
  \end{tcolorbox}

  \item A multiple-choice question (with its list of answer options) that refers to both the video and the image.
\end{enumerate}
}

\medskip
{\color{mymagenta}
Some additional assumptions to keep in mind:
\begin{itemize}[nosep,leftmargin=*]
    \item[--] Two furniture parts are \texttt{"}connected\texttt{"} if they are directly attached (physically in contact), like they
    would be in the final assembly.
    \item[--] Parts simply touching each other physically, but not in their final assembly position, are not considered \texttt{"}connected\texttt{"}.
    \item[--] While deciding the order of \texttt{"}connection\texttt{"} events between two parts, consider the order the order in which they
    were connected last as their actual connection, i.e., if two parts are connected, then disconnected and then connected again,
    the last connection is the one that matters.
    \item[--] In the course of the video, some parts may be attached physically to each other before they are fixed in their final assembly position
    by screwing, gluing, or otherwise securing them. In such cases, the parts are considered \texttt{"}connected\texttt{"} from the moment they are physically
    attached, not when they are fixed. Hence, when answering questions about the order of connections, consider the physical attachment
    as the connection event, even if the parts are not yet secured to each other.
\end{itemize}}

\medskip
{\color{mydarkblue}
Your responsibilities:
\begin{itemize}[nosep,leftmargin=*]
  \item[--] Watch and interpret the assembly steps in the video.
  \item[--] Examine the labeled image frame to understand part relationships.
  \item[--] Read the question and all answer choices.
  \item[--] Determine the correct option.
  \item[--] Respond **only** with a JSON object containing a single key, \texttt{"answer"}, whose value is the letter of your chosen option (e.g., \texttt{"A"}, \texttt{"B"}, \texttt{"C"}).
\end{itemize}

\medskip
**No additional text, explanations, or formatting \texttt{--} just the JSON string.**

Now answer the following question:
}
\end{tcolorbox}
\caption{\textbf{Task Instructions for the LVLMs.} We provide the models with task instructions that describe the input format, and additional information such as how we define \texttt{connection} events. When describing the inputs, we make adjustments for \track{} questions by mentioning that there are two visual prompts instead of one. The task instructions can be further broken down into 3 sections, describing the \textit{\bfseries \color{myorange} Inputs}, \textit{\bfseries \color{mymagenta} Assumptions}, and \textit{\bfseries \color{mydarkblue} Instructions} (colored accordingly above).}
\label{fig:taskdesc}
\vspace{-1.5em}
\end{figure*}

\begin{figure*}
\begin{tcolorbox}
[title=Task Instructions for Collage Prompts]
{\color{myorange}
You are a furniture-assembly expert. You are given a video input where each frame consists of two images side-by-side:
\begin{itemize}[nosep,leftmargin=*]
  \item[--] Right side: A frame from the furniture assembly video showing the assembly process.

  \item[--] Left side: A single labeled frame from the video on the right side, fixed for the entire video, displaying numeric IDs on each visible furniture part.
\end{itemize}

The left-side labeled frame remains constant throughout the video.

You will be asked multiple-choice questions referring to the assembly video and the labeled parts shown on the left.
}
\begin{tcolorbox}[
colback=gray!10,
colframe=gray!50,
boxrule=0.3pt,
left=1mm,
right=1mm,
top=0.5mm,
bottom=0.5mm,
title=\footnotesize For \track{} Questions,
coltitle=black,
fonttitle=\bfseries,
enhanced,
coltext=myorange
]
\footnotesize
You are a furniture-assembly expert. You are given a video input where each frame consists of three images side-by-side:
\begin{enumerate}[nosep,leftmargin=*]
  \item Right side: A frame from the furniture assembly video showing the assembly process.
  \item Left side: A labeled frame from the video on the right side, fixed for the entire video, displaying bright numeric IDs on each visible furniture part. Call this Image A.
  \item Center: Another labeled frame from the video on the right side, fixed for the entire video, also displaying bright numeric IDs on each visible furniture part. Call this Image B.
\end{enumerate}

Both Image A and Image B remain constant throughout the video.

Note: The numeric IDs in Image A and Image B are not necessarily the same; the same ID may refer to different parts in each image.
\medskip

You will be asked multiple-choice questions referring to the assembly video and the labeled parts shown on the left and central images.
\end{tcolorbox}

{\color{mymagenta} \huge$\mathbf{\cdot\cdot\cdot}$}

{\color{mydarkblue}
Your responsibilities:
\begin{enumerate}[nosep,leftmargin=*]
  \item Use the right-side video frames to observe the assembly steps.
  \item Use the fixed left-side labeled frame to identify and relate the furniture parts mentioned in the question.
  \begin{tcolorbox}[
    colback=gray!10,
    colframe=gray!50,
    boxrule=0.3pt,
    left=1mm,
    right=1mm,
    top=0.5mm,
    bottom=0.5mm,
    title=\footnotesize For \track{} Questions,
    coltitle=black,
    fonttitle=\bfseries,
    enhanced,
    coltext=mydarkblue
    ]
    \footnotesize
    \begin{enumerate}[nosep,leftmargin=*,start=2,label=\arabic*.,ref=\arabic*]
      \item Use the fixed labeled frames on the left side (Image A) and in the center (Image B) to identify and relate the furniture parts mentioned in the question.
    \end{enumerate}
    \end{tcolorbox}
  \item Carefully read the question and all answer choices.
  \item Select the correct answer based on the video and labeled parts.
  \item[--] Respond **only** with a JSON object containing a single key, \texttt{"answer"}, whose value is the letter of your chosen option (e.g., \texttt{"A"}, \texttt{"B"}, \texttt{"C"}).
\end{enumerate}

\medskip
**Do not include any explanations or additional text \texttt{--} reply with only the JSON string.**

Now answer the following question:}
\end{tcolorbox}
\caption{\textbf{Task Instructions for Collage Prompts.} \textit{\color{mymagenta} \bfseries Assumptions} remain the same (denoted by $\ldots$), but the \textit{\color{myorange} \bfseries Inputs} and \textit{\color{mydarkblue} \bfseries Instructions} are modified to describe the prompt format. In this case, the prompt image(s) are attached to the left of the video frame. For \track{} questions, the jumbled visual prompt is the second (middle) image in each frame.}
\label{fig:taskdesccollage}
\vspace{-1em}
\end{figure*}

\begin{figure*}
\begin{tcolorbox}
[title=Task Instructions for Concat Prompts]
{\color{myorange}
You are a furniture-assembly expert. You are given a video input of the assembly process of a furniture item.
The first frame of the video is a frame taken from the assembly process, highlighted with each visible furniture part, and displaying numeric IDs on each furniture part.

You will be asked multiple-choice questions referring to the assembly video and the labeled parts shown on the left.
}
\begin{tcolorbox}[
colback=gray!10,
colframe=gray!50,
boxrule=0.3pt,
left=1mm,
right=1mm,
top=0.5mm,
bottom=0.5mm,
title=\footnotesize For \track{} Questions,
coltitle=black,
fonttitle=\bfseries,
enhanced,
coltext=myorange
]
\footnotesize
You are a furniture-assembly expert. You are given a video input of the assembly process of a furniture item.
    The first two frames of the video are frames taken from the assembly process, highlighted with each visible furniture part, and
    displaying numeric IDs on each furniture part. We will refer to them as Image A (the first frame) and Image B (the second frame).

    Note: The numeric IDs in Image A and Image B are not necessarily the same; the same ID may refer to different parts in each image.

    You will be asked multiple-choice questions referring to the assembly video and the labeled parts shown in the first frame.
\end{tcolorbox}

{\color{mymagenta} \huge$\mathbf{\cdot\cdot\cdot}$}

{\color{mydarkblue}
Your responsibilities:
\begin{enumerate}[nosep,leftmargin=*]
  \item Use the video frames after the first frame to observe the assembly steps.
  \item Use the first labeled frame to identify and relate the furniture parts mentioned in the question.
  \begin{tcolorbox}[
    colback=gray!10,
    colframe=gray!50,
    boxrule=0.3pt,
    left=1mm,
    right=1mm,
    top=0.5mm,
    bottom=0.5mm,
    title=\footnotesize For \track{} Questions,
    coltitle=black,
    fonttitle=\bfseries,
    enhanced,
    coltext=mydarkblue,
    ]
    \footnotesize
    \begin{enumerate}[nosep,leftmargin=*,start=1,label=\arabic*.,ref=\arabic*]
      \item Use the video frames after the first two frames to observe the assembly steps.
      \item Use the first two labeled frames, i.e., Image A and Image B, to identify and relate the furniture parts mentioned in the question.
    \end{enumerate}
    \end{tcolorbox}
  \item Carefully read the question and all answer choices.
  \item Select the correct answer based on the video and labeled parts.
  \item[--] Respond **only** with a JSON object containing a single key, \texttt{"answer"}, whose value is the letter of your chosen option (e.g., \texttt{"A"}, \texttt{"B"}, \texttt{"C"}).
\end{enumerate}

\medskip
**Do not include any explanations or additional text \texttt{--} reply with only the JSON string.**

Now answer the following question:
}
\end{tcolorbox}
\caption{\textbf{Task Instructions for Concat Prompts.} \textit{\color{mymagenta} \bfseries Assumptions} remain the same (denoted by $\ldots$), but the \textit{\color{myorange} \bfseries Inputs} and \textit{\color{mydarkblue} \bfseries Instructions} are modified to describe the prompt format. In this case, the prompt image(s) are concatenated as the initial frames of the video. For \track{} questions, the jumbled visual prompt is the second frame in the video.}
\label{fig:taskdescconcat}
\end{figure*}

\paragraph{Final Prompt} The final prompt used in our evaluations has the following structure:\texttt{[Video Frames][Visual Prompt][Jumbled Visual Prompt][Task Instructions][Question]}.
The \texttt{[Jumbled Visual Prompt]} is exclusive to \textsc{Track} questions.

\paragraph{Human Evaluation} \Cref{fig:humanevalinst_combined} shows the instructions for the human participants who attempted our benchmark both the standard task and the image-only task~(\cref{sec:imageonly}). 
In the image-only task, we asked the participants to select an answer even if they were unsure.

\begin{figure*}[t]
    \centering

    \begin{minipage}[t]{0.48\textwidth}
        \begin{tcolorbox}
        \begin{enumerate}
            \item Each question consists of:
            \begin{enumerate}[nosep,leftmargin=*,label=\alph*.,ref=\alph*]
                \item A furniture assembly video.
                \item 1-2 visual prompts or frames extracted from the video, with certain parts shaded and labeled.
                \item An MCQ question with at most 4 options.
            \end{enumerate}
            \item Examine the labeled frame(s) and understand the relationships between the highlighted parts using the video.
            \item Read the question and then determine the correct option.
        \end{enumerate}
        \end{tcolorbox}
    \end{minipage}
    \hfill
    \begin{minipage}[t]{0.49\textwidth}
        \begin{tcolorbox}
        \small
        \begin{enumerate}
            \item Each question consists of:
            \begin{enumerate}[nosep,leftmargin=*,label=\alph*.,ref=\alph*]
                \item 1-2 images showing different stages in the assembly of a furniture item. Certain parts of the furniture will be shaded and labeled.
                \item An MCQ question about the furniture assembly process with at most 4 options.
            \end{enumerate}
            \item First, examine the labeled images(s) and understand the relationships between the highlighted parts.
            \item Read the question and then determine the correct option.
            \item Even if you cannot infer the answer from the question and the image alone, we request that you use your best judgment and select an option.
        \end{enumerate}
        \end{tcolorbox}
    \end{minipage}

    \caption{\textbf{Human Evaluation Instructions.}
    \textbf{Left:} Instructions for the standard task, where participants were provided the assembly video, visual prompt, and question text.
    \textbf{Right:} Instructions for the image-only task, they were provided only the visual prompts and question text.}
    \label{fig:humanevalinst_combined}
\end{figure*}

\begin{figure*}
    \centering
    \includegraphics[width=0.95\linewidth]{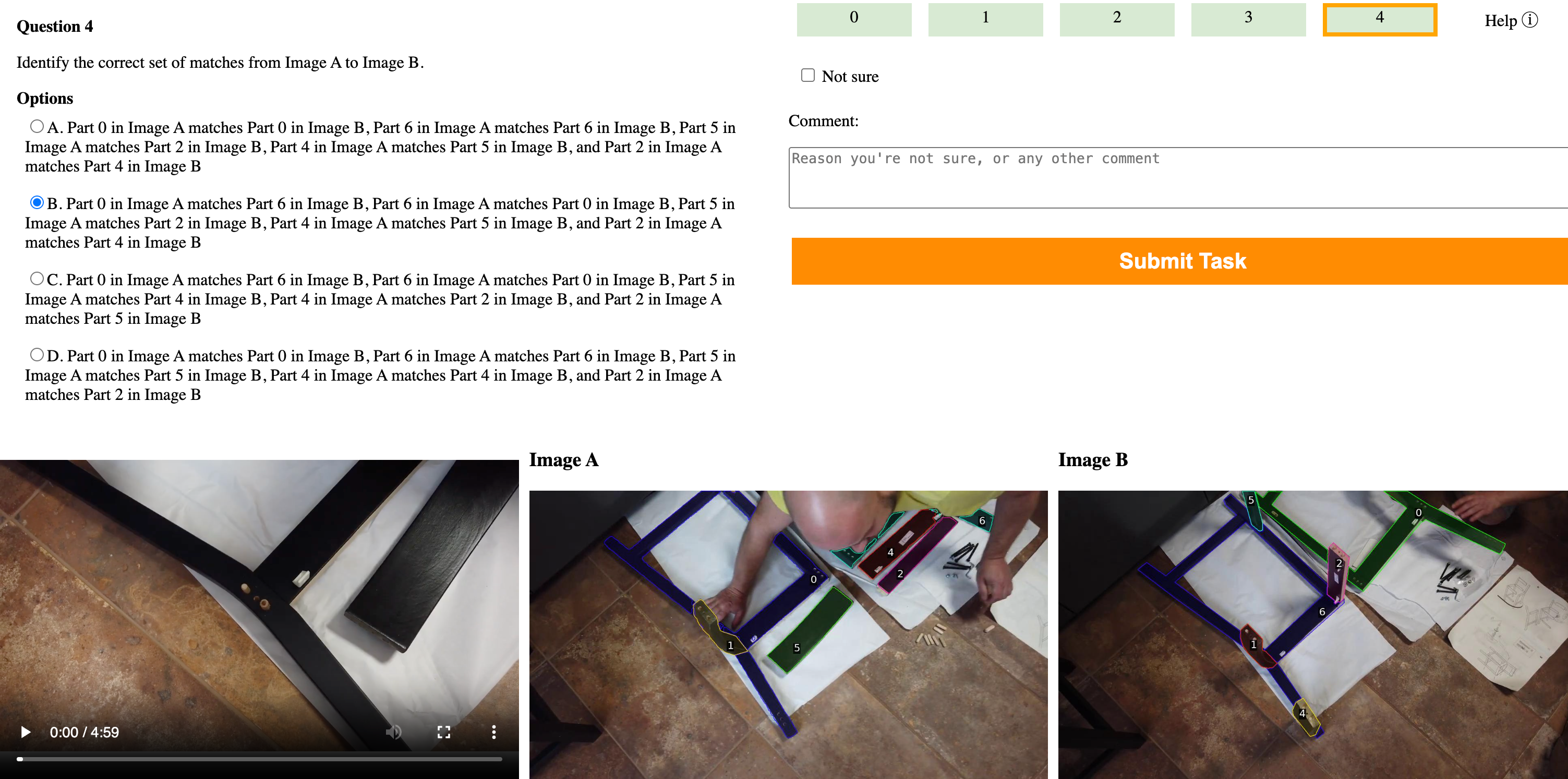}
    \caption{\textbf{Prolific Human Annotation Portal}. Users can scroll through the questions using the numbered buttons. Once they attempt all options, a button appears prompting them to submit their responses. They can access the instructions at any time using the ``\textit{Help}'' button.}
    \label{fig:annotui}
    \vspace{-1em}
\end{figure*}

In addition to the in-house human evaluation, we also conducted a broader human evaluation study on Prolific~\cite{prolific2025} with a smaller set of 186 questions.
Each participant was given a video and 4-5 questions related to the video. We collected at least 3 responses (up to 5) per question and computed the final answer using majority voting.
Figure~\ref{fig:annotui} shows the UI of the annotation tool that was shown to the participants.
We used a similar tool for our in-house annotations.
Results are shown in App.~\ref{app:results}.
We provide the instructions page that the Prolific participants received before beginning annotations on our project website: \href{https://flat-pack-bench.github.io/assets/prolific-instructions/}{\nolinkurl{flat-pack-bench.github.io/assets/prolific-instructions/}}.

\section{Additional Results}
\label{app:results}

\paragraph{Human Performance}
For the human performance study (\cref{sec:eval}), we recorded a micro-average unanimous response rate of 80\%, indicating that our questions are clear and consistently understood. 
Task-wise, \textsc{TOrd} attains the highest agreement (88\%), followed by \textsc{Mate} (86\%), and \textsc{Track} (77\%), and lastly \textsc{TLoc} (70\%).
Errors largely stemmed from confusing similar-looking parts over longer videos.

For the image-only human-performance results (\cref{tab:imageonly}), we observed a micro-average agreement rate of 25\%. Task-wise breakdowns for image-only variant were: 17\% for \textsc{TOrd}, 36\% for \textsc{TLoc}, 27\% for \textsc{Track}, and 20\% for \textsc{Mate}.

\begin{table}[t]
\centering
\caption{\textbf{Human evaluation summary on \flatpack.} We show the performance of Prolific and our in-house participants, along with some reference models on the same subset of questions. Agreement is measured by unanimous response rate.}
\renewcommand{\arraystretch}{1.15}
\setlength{\tabcolsep}{4pt}
\resizebox{\linewidth}{!}{
\begin{tabular}{l*{5}{r@{.}l}}
\toprule
\textbf{Cohort / Model} & \multicolumn{2}{c}{\textbf{Micro Avg.}} & \multicolumn{2}{c}{\textbf{\textsc{TOrd}}} & \multicolumn{2}{c}{\textbf{\textsc{TLoc}}} & \multicolumn{2}{c}{\textbf{\textsc{Track}}} & \multicolumn{2}{c}{\textbf{\textsc{Mate}}} \\
\rowcolor{sectionblue}
\sectiondividertwo{Human Evaluators}
Prolific Participants & 72 & 58 & 75 & 51 & 88 & 88 & 65 & 47 & 73 & 07 \\
Agreement & 27 & 95 & 28 & 57 & 37 & 04 & 21 & 43 & 38 & 46 \\
\hdashline
In-house Participants & { 98} & { 92} & { 100} & { 00} & { 100} & { 00} & { 97} & { 62} & { 100} & { 00} \\
Agreement & 84 & 94 & 91 & 83 & 70 & 37 & 85 & 71 & 84 & 61 \\
\sectiondividertwo{Reference Models}
InternVL3-78B & 44 & 09 & 44 & 90 & 51 & 85 & 41 & 67 & 42 & 31 \\
Qwen2.5-VL-72B & 45 & 70 & 46 & 94 & 29 & 63 & 48 & 81 & 50 & 00 \\
\bottomrule
\end{tabular}}
\label{tab:prolific}
\vspace{-1em}
\end{table}

The results of the broader human evaluation study discussed in \cref{app:evaldetails} are given in \cref{tab:prolific}.
We also compare these results with the performance of in-house participants on the same subset of questions.
As we can see, Prolific participants have a low agreement rate, but their performance is still significantly better than the reference models.
This shows that while the questions are generally solvable with high accuracy by humans, they certainly require high cognitive effort.

\paragraph{Zero-shot Results}
In~\cref{sec:eval} we stated that we evaluated all models on every combination of video type (key-frame, trimmed) and visual prompt type (Mixed-Media, Concat, and Collage). 
\cref{tab:maineval} showed the results for the best settings for each model.
We provide the complete results in~\cref{tab:mainevalfull}.
An interactive version of the table is also available on our project website\footnote{\projectsite[results/]}.
To obtain a more robust estimate of performance, we use \textit{bootstrapping} to compute the 95\% confidence interval (by sampling 50 videos with replacement for 100k trials).
If a setting is not shown, it was not evaluated due to cost constraints (e.g., GPT-5~\cite{openai2025gpt5} was assessed only on mixed-media prompts for key-frame videos) or because it does not support a prompt format (e.g., Mixed-Media prompts were not supported by some models, such as LLaVA-Video~\cite{zhang2024llavavideo}).

\paragraph{Correlation with Video Difficulty and Duration} To test whether model performance correlates with human-perceived difficulty, we manually annotated the videos on factors like the number of cut shots, the motion complexity of parts, etc. 
Figure \ref{fig:taskdescvideodiff} shows the instructions given to the manual annotators.
We collected two annotations per-video and averaged the scores to obtain a difficulty score for each video.
However, we found that for a standard open LVLM (Qwen2.5-VL 72B)  human-perceived difficulty scores do not predict model performance (Spearman’s $\rho$=-0.20, $p$=0.17).
Model performance also showed little correlation with video duration Spearman’s $\rho$=-0.21, $p$=0.13).

\begin{figure*}
\begin{tcolorbox}[title=Video Difficulty Annotation Instructions]
\small
You will be given a set of furniture assembly videos. Your task is to watch the video in its
entirety, and then answer some questions about the presence of certain artifacts or
characteristics in the video and how they affected your ability to understand the assembly
process.

Please judge difficulty with respect to answering fine-grained questions about part interactions,
ordering, and attachments (not merely recognizing the final assembled object).

\begin{enumerate}[nosep,leftmargin=*]
  \item \textbf{Zoomed-in/out cameras}: Are there frames where the camera is too
  zoomed-in/zoomed-out such that it becomes difficult to understand the assembly video?
  Answer based on overall impact on your understanding, not peak moments where this
  occurs. Please specify this difficulty on a scale of 1-3 where:
  \begin{enumerate}[nosep,leftmargin=*,label=\alph*.]
    \item 1 means zoom-in/zoom-out effects did not affect your understanding of the video
    at all
    \item 2 means you could not make out when 1-2 of the parts were attached due to the
    camera being too close or too far
    \item 3 means you could not understand $>50\%$ of the assembly process
  \end{enumerate}

  \item \textbf{Cut-shots}: Please assign a score to the video based on the number of cuts
  \begin{enumerate}[nosep,leftmargin=*,label=\alph*.]
    \item 0-1 cuts: 1
    \item 2-3 cuts: 2
    \item $>3$ cuts: 3
  \end{enumerate}

  \item \textbf{Out-of-frame rotations}: How difficult is it to understand out-of-frame (outside the
  camera’s view) rotations/attachments of parts in the video? Rate this on a scale of 1-3
  where:
  \begin{enumerate}[nosep,leftmargin=*,label=\alph*.]
    \item 1 means that the parts go out-of-frame for $<10$ frames and even when they do,
    the remain partially visible or their motion is very unambiguous
    \item 2 means that a noticeable number of parts ($>=20\%$) go completely out of frame
    for a few frames (10-50 frames), but you are still able to make out how they
    moved based on the other in-frame parts and the rest of the video
    \item 3 means that a noticeable number of parts ($>=20\%$) go out of frame for a
    significant duration of time ($>50$ frames), and you need to use guess work or your
    common-sense to understand how they have been manipulated when outside of
    the camera’s view.
  \end{enumerate}

  \item \textbf{Tracking or tracing the motion of moving parts}: We are trying to understand the amount
  of motion and and interactions between the parts in the video and whether it makes it
  difficult to understand which part was connected where. Rate the video from 1 to 3
  where:
  \begin{enumerate}[nosep,leftmargin=*,label=\alph*.]
    \item 1 means that the parts are interacted with minimally, typically only just before
    they are about to be attached.
    \item 2 means that some of the parts move around significantly, undergoing rotations
    even before they are about to be attached, but by 1-2 repeated viewings of the
    video, it is possible to understand the assembly process
    \item 3 means that the parts undergo complex motions and rotations during the video,
    and for some parts it is even difficult to be sure about its position in every frame.
    You had to use some guess work and elimination to track the parts through the
    full video.
  \end{enumerate}

  \item \textbf{Visual effects}: Does the video have visual effects like picture-in-picture, side-by-side
  visuals, etc.?
  \begin{enumerate}[nosep,leftmargin=*,label=\alph*.]
    \item 0 if no, 1 if yes
  \end{enumerate}

  \item \textbf{Text/Visual overlay}: Some of the videos you see might have text or pictures from the
  furniture assembly guide overlaid on the frames. Depending on how difficult did the
  overlaid content make it for you to understand the video either due to obscuring the
  visuals or providing ambiguous instructions, please assign a score from 1 to 3 where:
  \begin{enumerate}[nosep,leftmargin=*,label=\alph*.]
    \item 1 means that there was no overlaid context or even if was there, it does not
    obscure the visuals or make it difficult to understand the assembly process
    \item 2 means that even though overlaid content is present in some frames, it is
    translucent or does not obscure the visual content in the scene or adds to your
    understanding of the assembly process.
    \item 3 means that the overlaid content in some frames is opaque and completely
    obscures the visual content in the scene, making it difficult for you to understand
    the assembly and track the parts in the video
  \end{enumerate}

  \item \textbf{Initial orientation mismatch}: Does the assembly start in a flipped, upside-down, or
  non-canonical orientation relative to standard furniture assembly guides?
  \begin{enumerate}[nosep,leftmargin=*,label=\alph*.]
    \item 0: Canonical/upright orientation
    \item 1: Non-canonical (e.g., upside-down or rotated)
  \end{enumerate}
\end{enumerate}

\end{tcolorbox}
\vspace{0.95em}
\caption{\textbf{Instructions for annotation difficulty of videos.}  We asked annotators to rate the videos in our benchmark on different characteristics. The scores are then summed across characteristics and averaged across annotators to obtain a single score for the video.} 
\label{fig:taskdescvideodiff}
\end{figure*}

\section{Analysis Details}
\label{app:analysis}

\subsection{Self-Explanations}
\label{app:selfexpl}

\begin{figure}
    \centering
    \includegraphics[width=\linewidth]{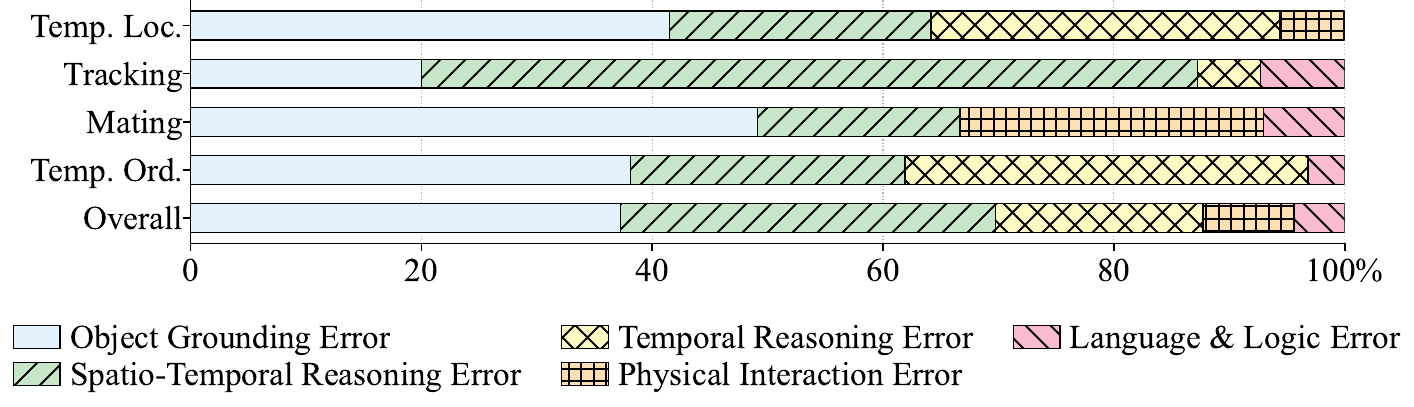}
    \caption{\textbf{Error Types for Rationales.} We provide the full quantitative distribution of error types for the rationales generated by Gemini 2.5 Pro, as discussed in~\cref{sec:selfexpl}.}
    \label{fig:selfexplerrordist}
\end{figure}

We have provided some additional examples of self-explanations, expanding upon the results discussed in~\cref{sec:selfexpl} on our project website.
We also provide the full breakdown of error categories for the generated rationales discussed in~\cref{sec:selfexpl} in~\cref{fig:selfexplerrordist}.

\subsection{Experimental Details for \tva{}}
\label{app:tvaresults}

We discussed an agentic baseline~\tva{} in~\cref{sec:tva} and reasons for its poor performance.
In this section, we provide the experimental details used to generate our results.

\paragraph{Contact-Reasoning Issues}
For our contact-reasoning experiment, we use the task instructions from the image-only prompting experiments in App.~\ref{app:imageonly}.
We used two versions of the question template:
\begin{enumerate}
    \item \textit{In this task, \texttt{"}connected\texttt{"} means the two parts are in direct physical contact, in the same way they will be when the furniture is fully assembled (not merely near each other or partially aligned). Based on the given image, are \textcolor{red}{\{query\_part1\}} and \textcolor{red}{\{query\_part2\}} connected?}
    \item \textit{Are \textcolor{red}{\{query\_part1\}} and \textcolor{red}{\{query\_part2\}} connected (physically in contact) in the shown image?}
\end{enumerate}

\Cref{tab:contact} shows the results. 
We observed poor performance across all settings, indicating that LVLMs struggle to understand even simpler concepts like physical contact.

\sisetup{
    table-number-alignment = center,
    round-mode = places,
    round-precision = 2
}

\begin{table}[t]
\centering
\caption{\textbf{Contact-Reasoning Results.} We show the performance of Qwen2.5-VL (32B \& 72B) across two question templates. The overall performance is quite poor across all settings.}
\renewcommand{\arraystretch}{1.2}
\renewcommand{\tabcolsep}{1.2mm}
\resizebox{\linewidth}{!}{
\begin{tabular}{l c
                S[table-format=2.2]
                S[table-format=2.2]
                S[table-format=2.2]}
\hline
\textbf{Model} & \textbf{Template} & 
\textbf{Micro Avg.} & \textbf{Yes} & \textbf{No} \\
\hline
Qwen2.5VL-32B & 1 & 62.60 & 41.20 & 84.00 \\
Qwen2.5VL-32B & 2 & 64.33       & 52.93 & 75.73 \\
Qwen2.5VL-72B & 1 & 58.80 & 33.33 & 84.26 \\
Qwen2.5VL-72B & 2 & 63.00 & 46.40        & 79.60 \\
\hline
\end{tabular}}
\label{tab:contact}
\vspace{-1em}
\end{table}

\paragraph{Tracking Issues}
We used SAM2~\cite{ravi2025sam} (\texttt{sam2.1\_hiera\_l}) for our tracking experiments.
We used trimmed videos for this experiment.
SAM2 cannot handle very long videos, so we implemented a chunk-wise propagation algorithm.
First, we split the video into 256-frame clips with a single frame overlap between adjacent clips.
Then, we first ran SAM2 on the clip containing the prompt frame.
After this, we propagate the masks forward in the video by using the masks predicted in the last frame of the initial clip to prompt the first frame in the next clip.
Similarly, we propagate the masks backward from the initial clip by using the masks predicted in the first frame of the initial clip to prompt the last frame in the preceding clip.
We continue this process until the entire video has been covered.
We subsampled frames by a factor of 4 to maintain a trade-off between compute constraints and temporal continuity.

\subsection{Linguistic Prompt Engineering}

This section provides additional details about the linguistic prompt engineering experiments discussed in~\cref{sec:promptengg}.
Initially, we added ``\textit{Please explain this answer step-by-step}'' to the task instruction. 
However, we found that the model was just paraphrasing its selected option.
Hence, we expanded this instruction to explicitly push the model to explain its reasoning. 
The results in~\cref{sec:promptengg}, and the qualitative results discussed in App.~\ref{app:selfexpl} are from this modified prompt.
We also amended the \textit{Instructions} portion of the task instruction to add an \texttt{explanation} key to the output JSON. 
\Cref{fig:taskdesccot} shows the complete task instructions.
In SC-CoT~\cite{wang2023selfconsistency}, for temperature sampling, we set the temperature to 0.7, and \texttt{top\_k} and \texttt{top\_p} set to 1 and 40, respectively~\cite{yang2024think}. 

\begin{figure*}[t]
\begin{tcolorbox}
[title=Task Instructions for CoT Prompting]
{\color{myorange}
You are a furniture-assembly expert. You are given:
\begin{enumerate}[nosep,leftmargin=*,label=\arabic*.,ref=\arabic*]
  \item A video of a furniture assembly in progress.

  \item A single labeled frame from the same video, displaying numeric IDs on each visible furniture part. 
  Call this a visual prompt.

  \begin{tcolorbox}[
    colback=gray!10,
    colframe=gray!50,
    boxrule=0.3pt,
    left=1mm,
    right=1mm,
    top=0.5mm,
    bottom=0.5mm,
    title=\footnotesize For \track{} Questions,
    coltitle=black,
    fonttitle=\bfseries,
    enhanced,
    coltext=myorange
  ]
  \footnotesize
  \begin{enumerate}[nosep,leftmargin=*,label=\arabic*.,ref=\arabic*,start=2]
      \item  Two labeled frames from the same video, called Image A and Image B, each displaying numeric IDs on visible furniture parts. Note: The numeric IDs in Image A and Image B are not necessarily the same; the same ID may refer to different parts in each image.
  \end{enumerate}
  \end{tcolorbox}

  \item A multiple-choice question (with its list of answer options) that refers to both the video and the image.
\end{enumerate}}

\medskip
{\color{mymagenta}
Some additional assumptions to keep in mind:
\begin{itemize}[nosep,leftmargin=*]
    \item[--] Two furniture parts are \texttt{"}connected\texttt{"} if they are directly attached (physically in contact), like they
    would be in the final assembly.
    \item[--] Parts simply touching each other physically, but not in their final assembly position, are not considered \texttt{"}connected\texttt{"}.
    \item[--] While deciding the order of \texttt{"}connection\texttt{"} events between two parts, consider the order the order in which they
    were connected last as their actual connection, i.e., if two parts are connected, then disconnected and then connected again,
    the last connection is the one that matters.
    \item[--] In the course of the video, some parts may be attached physically to each other before they are fixed in their final assembly position
    by screwing, gluing, or otherwise securing them. In such cases, the parts are considered \texttt{"}connected\texttt{"} from the moment they are physically
    attached, not when they are fixed. Hence, when answering questions about the order of connections, consider the physical attachment
    as the connection event, even if the parts are not yet secured to each other.
\end{itemize}}

\medskip
{\color{mydarkblue}
Your responsibilities:
\begin{itemize}[nosep,leftmargin=*]
  \item[--] Watch and interpret the assembly steps in the video.
  \item[--] Examine the labeled image frame to understand part relationships.

  \item[--] Read the question and all answer choices.
  \item[--] Determine the correct option.
  \item[--] Please explain your answer step-by-step. This explanation should contain your reasoning process, including how you analyzed the video and image to arrive at your conclusion. Do not simply restate the question or the answer choices.
  \item[--] Respond **only** with a JSON object containing two keys -
  \begin{enumerate}[nosep,leftmargin=*,label=\Alph*),ref=\Alph*)]
      \item \texttt{"answer"}, whose value is the letter of your chosen option (for example \texttt{"A", "B", "C"}, etc.).
      \item \texttt{"explanation"}, whose value is a step-by-step explanation of your answer. This must not be a restatement of the question or the answer choices, but rather a detailed reasoning process that leads to your answer. For example, describe how you analyzed the video and image to arrive at your conclusion or why certain options were eliminated. 
  \end{enumerate}
\end{itemize}

\medskip
**Do not include any explanations or additional text - reply with only the JSON string.**

Now answer the following question:}
\end{tcolorbox}
\caption{\textbf{Task Instructions for CoT.} We ask the model to explain its response, explicitly asking it not to paraphrase the correct option.}
\label{fig:taskdesccot}
\end{figure*}

\subsection{Image-only Prompts}
\label{app:imageonly}

In this section, we discuss the experimental details for the image-only prompts discussed in~\cref{sec:imageonly}.
\Cref{fig:taskdescimageonly} shows the task instruction used for our image-only experiments.
The remaining settings were the same as our main zero-shot evaluation in~\cref{sec:eval}.

\begin{figure*}[t]
\begin{tcolorbox}
[title=Task Instructions for Image-only Experiments]
{\color{myorange}
You are a furniture-assembly expert. You are given:
\begin{enumerate}[nosep,leftmargin=*,label=\arabic*.,ref=\arabic*]
  \item A still image showing a step in the assembly process of a furniture item, with the furniture parts shaded and labeled, displaying numeric IDs on each visible furniture part. Call this a visual prompt.
  \begin{tcolorbox}[
    colback=gray!10,
    colframe=gray!50,
    boxrule=0.3pt,
    left=1mm,
    right=1mm,
    top=0.5mm,
    bottom=0.5mm,
    title=\footnotesize For \track{} Questions,
    coltitle=black,
    fonttitle=\bfseries,
    enhanced,
    coltext=myorange
  ]
  \footnotesize
  \begin{enumerate}[nosep,leftmargin=*,label=\arabic*.,ref=\arabic*,start=2]
      \item  Two still images, Image A and Image B showing different steps in the assembly process of a furniture item, with the furniture parts shaded and labeled, displaying numeric IDs on each visible furniture part. 
  \end{enumerate}
  \end{tcolorbox}
  \item A multiple-choice question (with its list of answer options) that refers to the visual prompt.
\end{enumerate}}

\medskip
{\color{mymagenta}
Some additional assumptions to keep in mind:
\begin{itemize}[nosep,leftmargin=*]
    \item[--] Two furniture parts are \texttt{"}connected\texttt{"} if they are directly attached (physically in contact), like they
    would be in the final assembly.
    \item[--] Parts simply touching each other physically, but not in their final assembly position, are not considered \texttt{"}connected\texttt{"}.
\end{itemize}
}

\medskip

{\color{mydarkblue}
Your responsibilities:
\begin{itemize}[nosep,leftmargin=*]
    
  \item[--] Observe and interpret the assembly step shown in the image.
  \begin{tcolorbox}[
    colback=gray!10,
    colframe=gray!50,
    boxrule=0.3pt,
    left=1mm,
    right=1mm,
    top=0.5mm,
    bottom=0.5mm,
    title=\footnotesize For \track{} Questions,
    coltitle=black,
    fonttitle=\bfseries,
    enhanced,
    coltext=mydarkblue
  ]
  \footnotesize
  \begin{itemize}[nosep,leftmargin=*]
      \item[--]  Observe and interpret the assembly steps shown in the images.
  \end{itemize}
  \end{tcolorbox}
  \item[--] Examine the labeled image frames to understand part relationships.
  \item[--] Read the question and all answer choices.
  \item[--] Determine the correct option.
  \item[--] Respond **only** with a JSON object containing a single key, \texttt{`"answer"`}, whose value is the letter of your chosen option (e.g., \texttt{`"A"`, `"B"`, `"C"`}).
\end{itemize}

\medskip
**Do not include any explanations or additional text - reply with only the JSON string.**

Now answer the following question:
}
\end{tcolorbox}
\caption{\textbf{Image-only Task Instructions.} For image-only questions, the \textit{\textbf{\color{myorange} Inputs}} only consist of 1-2 prompt images. We remove any mention of the video from the \textit{\textbf{\color{mymagenta} Assumptions}} and the \textit{\textbf{\color{mydarkblue} Instructions}}.}
\label{fig:taskdescimageonly}
\end{figure*}

\section{Analysis on InternVL3} 
\label{app:internvlanalysis}

Here, we replicate the experiments discussed in \cref{sec:analysis} for the best model on our benchmark (\cref{tab:maineval}), InternVL3-78B~\cite{zhu2025internvl3exploringadvancedtraining}. 

\paragraph{Linguistic Prompt Engineering}
We follow the same prompts and generation settings as Qwen2.5-VL-72B (App.~\ref{app:analysis}). 
We use Concat prompts with Key-frame videos as they are the best choice for this model (See~\cref{tab:mainevalfull}).
\cref{tab:internvl_cot_comparison} shows our results.
ZS-CoT does lead to a minor improvement on \textsc{Track} but it does not improve overall performance, while SC-CoT leads to a significant decline in accuracy across all tasks.
Thus, similar to Qwen2.5-VL-72B, linguistic prompting strategies do not aid the spatio-temporal reasoning abilities of InternVL3-78B.

\begin{figure*}
    \centering
    \includegraphics[width=0.95\linewidth]{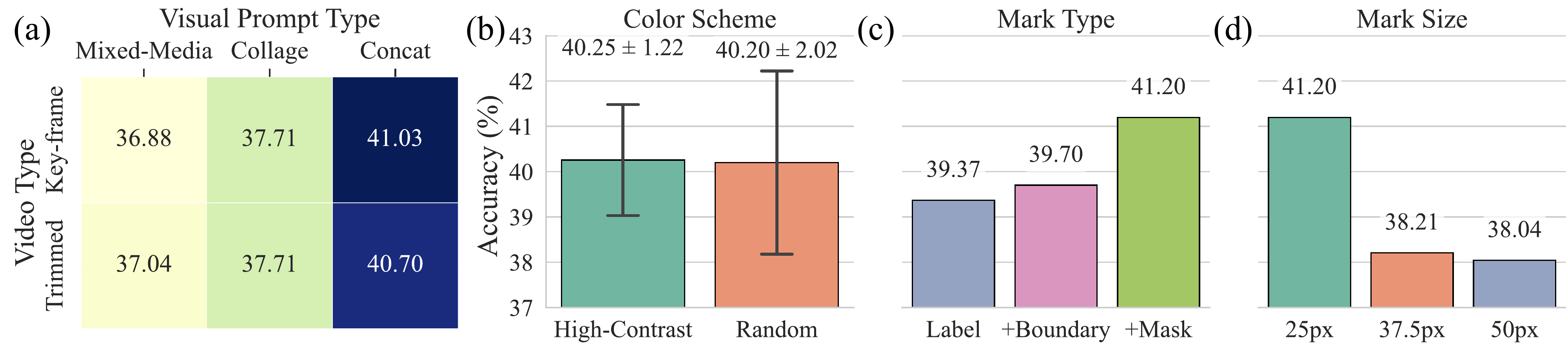}
    \caption{\textbf{Visual Data Ablation for InternVL3-78B.} InternVL3 perfoms better on Concat prompts with Key-frame videos, unlike Qwen2.5-VL 72B. However, when it comes to rendering the visual prompt images, both seem to follow similar trends}
    \label{fig:visualpromptablationinternvl}
\end{figure*}

\begin{table}[ht]
\centering
\small{
\caption{\textbf{Results of Lingustic Prompting Strategies for InternVL3-78B.} Similar to Qwen2.5-VL-72B, both ZS-CoT and SC-CoT fail to improve performance on \flatpack.}
\renewcommand{\arraystretch}{1.2}
\renewcommand{\tabcolsep}{1.2mm}
\resizebox{\linewidth}{!}{
\begin{tabular}{lccccc}
\toprule
 & \textbf{Micro Avg.} & \textbf{\textsc{TOrd}} & \textbf{\textsc{TLoc}} & \textbf{\textsc{Track}} & \textbf{\textsc{Mate}} \\
\midrule

-- & 41.03	& 43.87 & 39.80	& 42.02	& 34.48 \\ 
ZS-CoT & 41.03 & 41.29 & 37.86 & 44.75 & 33.33 \\
SC-CoT & 31.56 & 33.55 & 28.15 & 32.68 & 28.73 \\
\bottomrule
\end{tabular}}
\label{tab:internvl_cot_comparison}}
\vspace{-0.8em}
\end{table}

\paragraph{Visual Prompt Ablation}
\Cref{fig:visualpromptablationinternvl} shows the results of visual prompt ablation on InternVL3-78B. 
Firstly, we look at the different choices video and prompt types (\cref{fig:visualpromptablationinternvl}(a)). For InternVL3-78B the choice of visual prompt type does have a big impact as Concat prompt perform significantly better, unlike the mixed-media setting for Qwen2.5-VL-72B.
This is intuitive as InternVL3 was primarily trained on image-text and video-text sequence but not all three together (image-video-text sequences)~\cite{zhu2025internvl3exploringadvancedtraining}.
As before, the video type does not impact performance significantly.
With concat prompts and key-frame videos, we observe similar trends on color scheme, mark type and marker size to Qwen2.5-VL-72B.

\begin{table}[t]
\centering
\caption{\textbf{Image-only Prompt for InternVL3.} Performance of InternVL3-78B using image-only prompts, along with the change ($\Delta$) in performance from when the video is included in the prompt.}
\setlength{\tabcolsep}{5pt}
\renewcommand{\arraystretch}{1.18}
\renewcommand{\tabcolsep}{1.2mm}
\resizebox{\linewidth}{!}{
\begin{tabular}{lccccc}
\toprule
 & Micro Avg. & \textsc{TOrd} & \textsc{TLoc} & \textsc{Track} & \textsc{Mate} \\
\midrule
\textit{Video+Image}
& 41.03	& 43.87 & 39.81 & 42.02 & 34.48 \\
\hdashline
Image-only
& 27.57 & 32.19 & 36.89 & 20.23 & 26.44 \\
\quad$\Delta$
& \dneg{--13.46} & \dneg{-9.68} & \dneg{-2.91} & \dneg{--21.79} & \dneg{-8.05} \\
\hdashline
Shuffled Parts
& 29.35 \scriptsize{$\pm$ 1.25} & 35.48 \scriptsize{$\pm$ 1.94} & 39.16 \scriptsize{$\pm$ 2.02} & 19.20 \scriptsize{$\pm$ 0.98} & 36.78 \scriptsize{$\pm$ 1.15} \\
\quad$\Delta$
& \dneg{--11.68} 
& \dneg{--8.39}
& \dneg{--0.65}
& \dneg{--22.83}
& \dpos{+2.30} \\
\hdashline
\textbf{\textit{Human Perf.}} & 42.69 & 30.32 & 45.63 & 48.24 & 44.82 \\
\bottomrule
\end{tabular}}
\label{tab:imageonlyinternvl3}
\vspace{-1.5em}
\end{table}

\paragraph{Image-only Results}
On image-only prompts, InternVL3-78B shows clear degradation in performance across all tasks (See \cref{tab:imageonlyinternvl3}).
This suggests that it uses the video context more effectively, compared to Qwen2.5-VL-72B. 
This is also consistent with our observation that InternVL3-78B outperforms Qwen2.5-VL-72 on the main benchmark.
Despite this, the earlier trends of majority of the decline in performance stemming from \textsc{Track} persist, suggesting that even InternVL3-78B does not video context very effectively.

\medskip
\noindent
These results show that InternVL3-78B displays largely similar trends to Qwen2.5-VL-72B across all the questions we discussed in \cref{sec:analysis}.
This suggests that our conclusions are not specific to a single model and these findings may extend beyond any one model family or architecture.

\onecolumn
\begin{center}
\begingroup
\captionsetup{hypcap=false}
\captionof{table}{\textbf{Full Zero-shot Results on \flatpack.} In this table, we show the performance of all the models that we evaluated on each setting of video type and visual prompt type.}
\label{tab:mainevalfull}
\endgroup

\tablefirsthead{
  \toprule
  \textbf{Model} & \textbf{Prompt Type} & \textbf{Video Type} &
  \textbf{Micro Avg.} & \textbf{95\% CI} & \textbf{\textsc{TOrd}} &
  \textbf{\textsc{TLoc}} & \textbf{\textsc{Track}} &
  \textbf{\textsc{Mate}} \\
  \midrule
}
\tablehead{
  \multicolumn{9}{r}{\small\emph{(continued from previous page)}} \\
  \midrule
  \textbf{Model} & \textbf{Prompt Type} & \textbf{Video Type} &
  \textbf{Micro Avg.} & \textbf{95\% CI} & \textbf{\textsc{TOrd}} &
  \textbf{\textsc{TLoc}} & \textbf{\textsc{Track}} &
  \textbf{\textsc{Mate}} \\
  \midrule
}
\tabletail{
  \midrule
  \multicolumn{9}{r}{\small\emph{(table continues on next page)}} \\
}
\tablelasttail{
  \bottomrule
}
\begin{small}
\begin{supertabular}{lcccccccc}
Video-LLaVA-7B & Collage & Key-frame & 21.26 & [15.65, 26.88] & 21.94 & 30.10 & 10.51 & 41.38 \\
Video-LLaVA-7B & Collage & Trimmed & 20.76 & [15.37, 26.10] & 22.58 & 26.21 & 10.89 & 40.23 \\
Video-LLaVA-7B & Concat & Key-frame & 22.59 & [18.02, 27.11] & 23.87 & 32.04 & 12.45 & 39.08 \\
Video-LLaVA-7B & Concat & Trimmed & 21.59 & [16.80, 26.37] & 23.23 & 29.13 & 12.06 & 37.93 \\
Video-LLaVA-7B & Mixed-Media & Key-frame & 23.75 & [18.12, 29.40] & 21.29 & 35.92 & 10.89 & 51.72 \\
Video-LLaVA-7B & Mixed-Media & Trimmed & 23.59 & [18.00, 29.23] & 21.94 & 34.95 & 10.51 & 51.72 \\
InternVL3-14B & Collage & Key-frame & 37.71 & [32.02, 43.41] & 42.58 & 21.36 & 37.74 & 48.28 \\
InternVL3-14B & Collage & Trimmed & 37.04 & [32.58, 41.29] & 40.00 & 28.16 & 35.41 & 47.13 \\
InternVL3-14B & Concat & Key-frame & 34.88 & [30.46, 38.87] & 37.42 & 31.07 & 33.46 & 39.08 \\
InternVL3-14B & Concat & Trimmed & 33.55 & [27.98, 38.47] & 38.06 & 30.10 & 31.13 & 36.78 \\
InternVL3-14B & Mixed-Media & Key-frame & 35.05 & [31.30, 38.59] & 39.35 & 37.86 & 31.13 & 35.63 \\
InternVL3-14B & Mixed-Media & Trimmed & 35.88 & [31.88, 40.19] & 38.71 & 39.81 & 31.52 & 39.08 \\
InternVL3-38B & Collage & Key-frame & 34.72 & [29.82, 39.26] & 37.42 & 33.98 & 26.85 & 54.02 \\
InternVL3-38B & Collage & Trimmed & 35.05 & [30.20, 39.41] & 38.06 & 33.98 & 26.85 & 55.17 \\
InternVL3-38B & Concat & Key-frame & 36.05 & [31.38, 40.06] & 42.58 & 37.86 & 25.68 & 52.87 \\
InternVL3-38B & Concat & Trimmed & 35.22 & [30.60, 39.23] & 37.42 & 41.75 & 26.07 & 50.57 \\
InternVL3-38B & Mixed-Media & Key-frame & 33.06 & [28.49, 37.91] & 45.81 & 47.57 & 15.95 & 43.68 \\
InternVL3-38B & Mixed-Media & Trimmed & 31.73 & [26.77, 36.44] & 45.16 & 41.75 & 15.56 & 43.68 \\
InternVL3-78B & Collage & Key-frame & 37.71 & [32.31, 42.72] & 32.90 & 31.07 & 43.97 & 35.63 \\
InternVL3-78B & Collage & Trimmed & 37.71 & [32.22, 42.73] & 33.55 & 33.01 & 42.80 & 35.63 \\
InternVL3-78B & Concat & Key-frame & 41.03 & [36.21, 45.64] & 43.87 & 39.81 & 42.02 & 34.48 \\
InternVL3-78B & Concat & Trimmed & 40.70 & [35.81, 45.00] & 44.52 & 37.86 & 42.02 & 33.33 \\
InternVL3-78B & Mixed-Media & Key-frame & 36.88 & [32.66, 41.27] & 33.55 & 44.66 & 36.96 & 33.33 \\
InternVL3-78B & Mixed-Media & Trimmed & 37.04 & [33.33, 40.92] & 36.77 & 44.66 & 36.96 & 28.74 \\
Qwen2.5-VL-32B & Collage & Key-frame & 35.88 & [31.18, 40.38] & 34.84 & 29.13 & 33.07 & 54.02 \\
Qwen2.5-VL-32B & Collage & Trimmed & 32.72 & [28.40, 36.92] & 34.19 & 24.27 & 28.02 & 54.02 \\
Qwen2.5-VL-32B & Concat & Key-frame & 28.57 & [23.86, 33.06] & 32.90 & 30.10 & 18.68 & 48.28 \\
Qwen2.5-VL-32B & Concat & Trimmed & 27.57 & [22.95, 32.23] & 32.26 & 28.16 & 16.73 & 50.57 \\
Qwen2.5-VL-32B & Mixed-Media & Key-frame & 32.72 & [28.27, 37.16] & 38.71 & 33.98 & 24.51 & 44.83 \\
Qwen2.5-VL-32B & Mixed-Media & Trimmed & 34.72 & [30.40, 38.77] & 38.71 & 42.72 & 26.07 & 43.68 \\
Qwen2.5-VL-72B & Collage & Key-frame & 37.87 & [31.79, 43.22] & 33.55 & 23.30 & 42.41 & 49.43 \\
Qwen2.5-VL-72B & Collage & Trimmed & 34.88 & [29.23, 40.03] & 38.06 & 25.24 & 32.68 & 47.13 \\
Qwen2.5-VL-72B & Concat & Key-frame & 33.06 & [26.84, 38.73] & 35.48 & 26.21 & 28.40 & 50.57 \\
Qwen2.5-VL-72B & Concat & Trimmed & 32.72 & [26.72, 38.12] & 34.19 & 24.27 & 29.18 & 50.57 \\
Qwen2.5-VL-72B & Mixed-Media & Key-frame & 40.20 & [35.22, 44.60] & 40.65 & 30.10 & 45.53 & 35.63 \\
Qwen2.5-VL-72B & Mixed-Media & Trimmed & 40.37 & [34.81, 45.29] & 41.29 & 30.10 & 45.14 & 36.78 \\
Qwen2.5-VL-7B & Collage & Key-frame & 25.58 & [22.24, 28.95] & 30.32 & 23.30 & 18.68 & 40.23 \\
Qwen2.5-VL-7B & Collage & Trimmed & 25.08 & [21.44, 28.79] & 30.32 & 24.27 & 17.90 & 37.93 \\
Qwen2.5-VL-7B & Concat & Key-frame & 28.74 & [25.07, 32.39] & 30.97 & 21.36 & 25.68 & 42.53 \\
Qwen2.5-VL-7B & Concat & Trimmed & 29.24 & [25.13, 33.10] & 29.68 & 22.33 & 26.07 & 45.98 \\
Qwen2.5-VL-7B & Mixed-Media & Key-frame & 30.23 & [24.83, 35.38] & 27.10 & 18.45 & 33.07 & 41.38 \\
Qwen2.5-VL-7B & Mixed-Media & Trimmed & 29.57 & [25.27, 33.67] & 30.97 & 21.36 & 28.02 & 41.38 \\
Qwen3-VL-235B-A22B & Collage & Key-frame & 37.21 & [32.77, 41.59] & 37.42 & 25.24 & 39.69 & 43.68 \\
Qwen3-VL-235B-A22B & Collage & Trimmed & 36.05 & [29.68, 42.02] & 36.77 & 23.30 & 35.41 & 51.72 \\
Qwen3-VL-235B-A22B & Concat & Key-frame & 34.88 & [30.31, 39.31] & 38.06 & 33.01 & 32.30 & 39.08 \\
Qwen3-VL-235B-A22B & Concat & Trimmed & 34.72 & [28.65, 40.31] & 36.13 & 24.27 & 32.30 & 51.72 \\
Qwen3-VL-235B-A22B & Mixed-Media & Key-frame & 33.55 & [28.60, 39.05] & 37.42 & 38.83 & 28.79 & 34.48 \\
Qwen3-VL-235B-A22B & Mixed-Media & Trimmed & 32.23 & [27.85, 36.27] & 35.48 & 40.78 & 26.07 & 34.48 \\
Qwen3-VL-30B-A3B & Collage & Key-frame & 35.38 & [30.74, 40.11] & 33.55 & 24.27 & 38.52 & 42.53 \\
Qwen3-VL-30B-A3B & Collage & Trimmed & 35.22 & [29.94, 40.36] & 31.61 & 22.33 & 36.96 & 51.72 \\
Qwen3-VL-30B-A3B & Concat & Key-frame & 36.05 & [30.65, 41.31] & 34.84 & 20.39 & 42.02 & 39.08 \\
Qwen3-VL-30B-A3B & Concat & Trimmed & 36.71 & [30.96, 42.17] & 30.32 & 22.33 & 42.02 & 49.43 \\
Qwen3-VL-30B-A3B & Mixed-Media & Key-frame & 36.54 & [32.25, 40.82] & 35.48 & 35.92 & 36.19 & 40.23 \\
Qwen3-VL-30B-A3B & Mixed-Media & Trimmed & 36.38 & [32.22, 40.76] & 36.13 & 32.04 & 38.91 & 34.48 \\
Qwen3-VL-32B & Collage & Key-frame & 35.05 & [30.37, 39.16] & 36.13 & 32.04 & 33.07 & 42.53 \\
Qwen3-VL-32B & Collage & Trimmed & 34.05 & [28.57, 39.24] & 40.65 & 25.24 & 31.13 & 41.38 \\
Qwen3-VL-32B & Concat & Key-frame & 32.06 & [26.96, 36.86] & 36.13 & 28.16 & 28.02 & 41.38 \\
Qwen3-VL-32B & Concat & Trimmed & 33.06 & [27.63, 38.35] & 35.48 & 24.27 & 31.13 & 44.83 \\
Qwen3-VL-32B & Mixed-Media & Key-frame & 37.71 & [33.09, 42.11] & 38.71 & 46.60 & 31.91 & 42.53 \\
Qwen3-VL-32B & Mixed-Media & Trimmed & 35.71 & [31.51, 39.88] & 38.71 & 44.66 & 28.40 & 41.38 \\
Qwen3-VL-32B-Think & Collage & Key-frame & 31.56 & [26.42, 36.20] & 36.13 & 26.21 & 27.63 & 41.38 \\
Qwen3-VL-32B-Think & Collage & Trimmed & 40.03 & [33.82, 45.50] & 38.71 & 22.33 & 45.53 & 47.13 \\
Qwen3-VL-32B-Think & Concat & Key-frame & 24.25 & [20.55, 28.13] & 34.19 & 21.36 & 18.29 & 27.59 \\
Qwen3-VL-32B-Think & Concat & Trimmed & 34.88 & [30.09, 39.29] & 35.48 & 25.24 & 33.85 & 48.28 \\
Qwen3-VL-32B-Think & Mixed-Media & Key-frame & 34.72 & [29.98, 39.07] & 41.29 & 33.01 & 30.35 & 37.93 \\
Qwen3-VL-32B-Think & Mixed-Media & Trimmed & 35.38 & [30.30, 39.91] & 40.00 & 33.98 & 32.30 & 37.93 \\
Qwen3-VL-4B & Collage & Key-frame & 35.55 & [30.69, 40.10] & 32.26 & 33.98 & 31.13 & 56.32 \\
Qwen3-VL-4B & Collage & Trimmed & 35.38 & [31.03, 39.52] & 31.61 & 31.07 & 33.85 & 51.72 \\
Qwen3-VL-4B & Concat & Key-frame & 32.56 & [27.21, 37.92] & 32.90 & 25.24 & 29.96 & 48.28 \\
Qwen3-VL-4B & Concat & Trimmed & 36.54 & [31.72, 41.20] & 34.19 & 33.01 & 32.68 & 56.32 \\
Qwen3-VL-4B & Mixed-Media & Key-frame & 34.39 & [29.45, 39.21] & 32.90 & 30.10 & 32.68 & 47.13 \\
Qwen3-VL-4B & Mixed-Media & Trimmed & 34.55 & [29.46, 39.43] & 30.97 & 29.13 & 34.24 & 48.28 \\
Qwen3-VL-4B-Think & Collage & Key-frame & 27.57 & [21.96, 32.45] & 30.32 & 25.24 & 19.46 & 49.43 \\
Qwen3-VL-4B-Think & Collage & Trimmed & 31.73 & [26.13, 37.16] & 34.84 & 25.24 & 26.85 & 48.28 \\
Qwen3-VL-4B-Think & Concat & Key-frame & 32.23 & [27.05, 36.64] & 32.90 & 29.13 & 29.96 & 41.38 \\
Qwen3-VL-4B-Think & Concat & Trimmed & 37.21 & [32.11, 41.78] & 31.61 & 25.24 & 37.74 & 59.77 \\
Qwen3-VL-4B-Think & Mixed-Media & Key-frame & 27.41 & [21.65, 32.31] & 29.03 & 25.24 & 23.35 & 39.08 \\
Qwen3-VL-4B-Think & Mixed-Media & Trimmed & 28.90 & [23.72, 33.42] & 28.39 & 31.07 & 23.74 & 42.53 \\
Qwen3-VL-8B & Collage & Key-frame & 28.57 & [24.44, 32.87] & 36.13 & 32.04 & 20.62 & 34.48 \\
Qwen3-VL-8B & Collage & Trimmed & 25.58 & [21.02, 31.00] & 31.61 & 23.30 & 17.90 & 40.23 \\
Qwen3-VL-8B & Concat & Key-frame & 28.07 & [23.31, 33.39] & 32.26 & 27.18 & 22.57 & 37.93 \\
Qwen3-VL-8B & Concat & Trimmed & 29.40 & [24.41, 34.18] & 33.55 & 23.30 & 26.46 & 37.93 \\
Qwen3-VL-8B & Mixed-Media & Key-frame & 33.72 & [29.31, 38.21] & 36.13 & 30.10 & 33.85 & 33.33 \\
Qwen3-VL-8B & Mixed-Media & Trimmed & 31.73 & [26.70, 37.28] & 34.19 & 31.07 & 31.91 & 27.59 \\
Qwen3-VL-8B-Think & Collage & Key-frame & 28.41 & [23.60, 32.55] & 30.97 & 27.18 & 21.40 & 45.98 \\
Qwen3-VL-8B-Think & Collage & Trimmed & 29.07 & [24.14, 33.70] & 35.48 & 26.21 & 20.23 & 47.13 \\
Qwen3-VL-8B-Think & Concat & Key-frame & 25.25 & [20.37, 29.74] & 32.90 & 31.07 & 14.79 & 35.63 \\
Qwen3-VL-8B-Think & Concat & Trimmed & 27.41 & [22.83, 31.52] & 34.19 & 21.36 & 19.07 & 47.13 \\
Qwen3-VL-8B-Think & Mixed-Media & Key-frame & 26.58 & [22.00, 30.82] & 34.84 & 26.21 & 19.46 & 33.33 \\
Qwen3-VL-8B-Think & Mixed-Media & Trimmed & 31.73 & [27.12, 35.80] & 34.19 & 33.01 & 25.29 & 44.83 \\
Perception-LM-1B & Collage & Key-frame & 27.74 & [22.85, 31.96] & 28.39 & 26.21 & 25.29 & 35.63 \\
Perception-LM-1B & Collage & Trimmed & 27.41 & [22.92, 31.37] & 27.10 & 25.24 & 26.07 & 34.48 \\
Perception-LM-1B & Concat & Key-frame & 27.57 & [23.48, 31.23] & 24.52 & 28.16 & 26.46 & 35.63 \\
Perception-LM-1B & Concat & Trimmed & 27.41 & [23.44, 30.88] & 23.87 & 27.18 & 26.85 & 35.63 \\
Perception-LM-3B & Collage & Key-frame & 29.40 & [24.95, 33.63] & 27.74 & 32.04 & 26.46 & 37.93 \\
Perception-LM-3B & Collage & Trimmed & 31.40 & [27.15, 35.26] & 28.39 & 32.04 & 29.96 & 40.23 \\
Perception-LM-3B & Concat & Key-frame & 29.40 & [24.38, 34.12] & 29.03 & 33.98 & 26.46 & 33.33 \\
Perception-LM-3B & Concat & Trimmed & 28.90 & [24.13, 33.13] & 28.39 & 31.07 & 27.24 & 32.18 \\
Perception-LM-8B & Collage & Key-frame & 35.22 & [29.53, 40.28] & 25.16 & 26.21 & 44.75 & 35.63 \\
Perception-LM-8B & Collage & Trimmed & 35.38 & [29.29, 41.22] & 26.45 & 26.21 & 44.75 & 34.48 \\
Perception-LM-8B & Concat & Key-frame & 29.90 & [24.55, 34.66] & 20.65 & 19.42 & 38.13 & 34.48 \\
Perception-LM-8B & Concat & Trimmed & 30.90 & [25.77, 35.68] & 25.16 & 22.33 & 37.35 & 32.18 \\
Gemini 2.5 Flash & Collage & Key-frame & 23.26 & [19.93, 26.29] & 25.81 & 33.98 & 12.06 & 39.08 \\
Gemini 2.5 Flash & Collage & Trimmed & 18.77 & [15.23, 21.91] & 19.35 & 30.10 & 10.89 & 27.59 \\
Gemini 2.5 Flash & Concat & Key-frame & 27.41 & [24.01, 31.05] & 32.90 & 42.72 & 13.23 & 41.38 \\
Gemini 2.5 Flash & Concat & Trimmed & 23.09 & [18.94, 27.76] & 27.10 & 33.01 & 11.28 & 39.08 \\
Gemini 2.5 Flash & Mixed-Media & Key-frame & 31.06 & [27.12, 35.45] & 31.61 & 41.75 & 23.35 & 40.23 \\
Gemini 2.5 Flash & Mixed-Media & Trimmed & 26.25 & [21.55, 30.63] & 27.74 & 33.01 & 20.23 & 33.33 \\
Gemini 2.5 Pro & Concat & Key-frame & 32.22 & [28.92, 35.85] & 39.35 & 51.45 & 18.67 & 36.78 \\
Gemini 2.5 Pro & Mixed-Media & Key-frame & 33.72 & [30.16, 37.11] & 40.65 & 44.66 & 23.35 & 39.08 \\
Gemini 3.1 Pro & Concat & Key-frame & 32.89 & [28.53, 37.13] & 34.84 & 43.69 & 21.79 & 49.42 \\
GPT-5 & Mixed-Media & Key-frame & 37.71 & [32.79, 42.89] & 40.65 & 53.40 & 25.68 & 49.43 \\
LLaVA-Next-Vid-34B & Collage & Key-frame & 27.91 & [23.24, 32.71] & 29.68 & 28.16 & 28.02 & 24.14 \\
LLaVA-Next-Vid-34B & Collage & Trimmed & 26.91 & [22.58, 31.38] & 30.97 & 26.21 & 26.07 & 22.99 \\
LLaVA-Next-Vid-34B & Concat & Key-frame & 28.07 & [22.98, 33.44] & 31.61 & 24.27 & 29.57 & 21.84 \\
LLaVA-Next-Vid-34B & Concat & Trimmed & 29.07 & [24.53, 33.67] & 32.26 & 24.27 & 30.74 & 24.14 \\
LLaVA-Next-Vid-34B & Mixed-Media & Key-frame & 28.90 & [24.44, 33.08] & 27.74 & 26.21 & 31.52 & 26.44 \\
LLaVA-Next-Vid-34B & Mixed-Media & Trimmed & 30.40 & [25.87, 34.69] & 30.32 & 24.27 & 32.68 & 31.03 \\
LLaVA-Next-Vid-7B & Collage & Key-frame & 23.26 & [20.14, 26.46] & 29.03 & 26.21 & 15.56 & 32.18 \\
LLaVA-Next-Vid-7B & Collage & Trimmed & 25.08 & [22.12, 28.39] & 33.55 & 24.27 & 16.73 & 35.63 \\
LLaVA-Next-Vid-7B & Concat & Key-frame & 23.75 & [20.46, 26.86] & 29.68 & 29.13 & 10.89 & 44.83 \\
LLaVA-Next-Vid-7B & Concat & Trimmed & 24.25 & [21.23, 27.14] & 30.97 & 29.13 & 11.67 & 43.68 \\
LLaVA-Next-Vid-7B & Mixed-Media & Key-frame & 24.58 & [21.50, 27.37] & 26.45 & 26.21 & 17.12 & 41.38 \\
LLaVA-Next-Vid-7B & Mixed-Media & Trimmed & 24.42 & [21.66, 27.24] & 27.10 & 26.21 & 15.18 & 44.83 \\
LlaVA-OneVision-72B & Collage & Key-frame & 37.87 & [32.87, 42.55] & 34.84 & 26.21 & 37.74 & 57.47 \\
LlaVA-OneVision-72B & Collage & Trimmed & 38.37 & [32.97, 43.26] & 35.48 & 25.24 & 38.91 & 57.47 \\
LlaVA-OneVision-72B & Concat & Key-frame & 36.05 & [31.45, 40.09] & 35.48 & 27.18 & 33.46 & 55.17 \\
LlaVA-OneVision-72B & Concat & Trimmed & 36.05 & [31.32, 40.33] & 35.48 & 25.24 & 34.24 & 55.17 \\
LlaVA-OneVision-7B & Collage & Key-frame & 29.07 & [25.60, 32.33] & 23.87 & 25.24 & 30.35 & 39.08 \\
LlaVA-OneVision-7B & Collage & Trimmed & 28.24 & [24.46, 31.39] & 23.87 & 20.39 & 29.18 & 42.53 \\
LlaVA-OneVision-7B & Concat & Key-frame & 32.39 & [28.09, 36.16] & 25.16 & 31.07 & 33.07 & 44.83 \\
LlaVA-OneVision-7B & Concat & Trimmed & 32.89 & [28.76, 36.48] & 26.45 & 30.10 & 34.24 & 43.68 \\
LlaVA-OneVision-7B & Mixed-Media & Key-frame & 31.23 & [26.83, 35.14] & 27.74 & 29.13 & 33.46 & 33.33 \\
LlaVA-OneVision-7B & Mixed-Media & Trimmed & 31.56 & [27.64, 35.13] & 28.39 & 26.21 & 34.63 & 34.48 \\
LLaVA-Video-72B & Collage & Key-frame & 37.54 & [31.93, 42.53] & 36.77 & 27.18 & 35.80 & 56.32 \\
LLaVA-Video-72B & Collage & Trimmed & 34.39 & [29.64, 38.63] & 33.55 & 21.36 & 35.02 & 49.43 \\
LLaVA-Video-72B & Concat & Key-frame & 34.05 & [28.85, 38.97] & 38.71 & 27.18 & 27.24 & 54.02 \\
LLaVA-Video-72B & Concat & Trimmed & 33.39 & [28.82, 37.39] & 40.00 & 25.24 & 25.68 & 54.02 \\
LLaVA-Video-7B & Collage & Key-frame & 28.57 & [23.85, 33.06] & 23.23 & 28.16 & 23.74 & 52.87 \\
LLaVA-Video-7B & Collage & Trimmed & 28.57 & [24.88, 32.70] & 26.45 & 28.16 & 24.12 & 45.98 \\
LLaVA-Video-7B & Concat & Key-frame & 30.73 & [25.33, 36.06] & 30.97 & 24.27 & 25.68 & 52.87 \\
LLaVA-Video-7B & Concat & Trimmed & 27.74 & [22.11, 33.33] & 24.52 & 21.36 & 26.85 & 43.68 \\
ArrowRL-7B & Collage & Key-frame & 28.90 & [24.40, 33.02] & 36.77 & 26.21 & 20.62 & 42.53 \\
ArrowRL-7B & Collage & Trimmed & 24.92 & [20.51, 29.29] & 27.74 & 21.36 & 20.23 & 37.93 \\
ArrowRL-7B & Concat & Key-frame & 30.07 & [26.68, 33.14] & 34.84 & 24.27 & 25.68 & 41.38 \\
ArrowRL-7B & Concat & Trimmed & 30.56 & [26.71, 34.19] & 30.97 & 24.27 & 29.18 & 41.38 \\
ArrowRL-7B & Mixed-Media & Key-frame & 29.90 & [25.36, 34.35] & 33.55 & 23.30 & 28.40 & 35.63 \\
ArrowRL-7B & Mixed-Media & Trimmed & 30.40 & [25.63, 35.18] & 30.97 & 28.16 & 28.02 & 39.08 \\
Gemini 2.5 Pro + GenS & Collage & Key-frame & 25.58 & [21.70, 29.67] & 33.55 & 32.04 & 13.23 & 40.23 \\
VideoRefer & --- & Key-frame & 28.57 & [28.50, 33.33] & 32.90 & 30.10 & 17.51 & 51.72 \\
\end{supertabular}
\end{small}
\end{center}
\twocolumn

\end{document}